%% file: 00_main_arr.tex
\pdfoutput=1
\documentclass[11pt]{article}

\usepackage[final]{acl}
\usepackage{times}
\usepackage{latexsym}

\usepackage[T1]{fontenc}
\usepackage[utf8]{inputenc}
\usepackage{microtype}
\usepackage{inconsolata}
\usepackage{graphicx}

\usepackage{booktabs}
\usepackage{url}
\usepackage{subcaption}
\usepackage{multirow}
\usepackage{tabularx}
\usepackage[most]{tcolorbox}
\usepackage{enumitem}
\usepackage{numprint} 

\usepackage{amsmath}
\usepackage{amssymb}
\usepackage{mathtools}
\usepackage{amsthm}

\usepackage{algorithm}
\usepackage{algpseudocode}

\newcommand\model{\textsc{AuGLM\ }}
\newcommand\modelns{\textsc{AuGLM}}

\title{How to Make LMs Strong Node Classifiers?}

\author{
  \textbf{Zhe Xu}$^{12}$, \textbf{Kaveh Hassani}$^{1}$, \textbf{Si Zhang}$^{1}$, \textbf{Hanqing Zeng}$^{1}$, \textbf{Michihiro Yasunaga}$^{1}$, \\
  \textbf{Limei Wang}$^{1}$, \textbf{Dongqi Fu}$^{1}$, \textbf{Ning Yao}$^{1}$, \textbf{Bo Long}$^{1}$, \textbf{Hanghang Tong}$^2$ \\
  $^{1}$Meta, $^{2}$University of Illinois Urbana-Champaign \\
  \texttt{\{zhexu,kavehhassani,sizhang\}@meta.com, htong@illinois.edu}
}

\begin{document}
\maketitle

\input{Files/00Abstract}
\input{Files/01Intro}
\input{Files/05RelatedWork}
\input{Files/02Prelim}
\input{Files/03Method}
\input{Files/04Exp}
\input{Files/06Conclusion}


\section{Limitations}
\label{sec: limitation and future work}
One limitation of this work is the need for manual definition of prompt templates in Table~\ref{tab: templates}. A promising direction for future research is to develop methods for automatically searching for optimal templates in a data-driven manner. Another limitation is the requirement for pretraining a GNN $\psi$ on each dataset, which stems from the inherent challenges of language models in understanding graph data and limits our model to be adapted to zero-shot learning scenarios. Addressing this limitation by developing more powerful language models capable of handling graph data is a challenging yet impactful area of future work, which can lead to instruction-tuning only, highly generalizable graph foundation models.

\section{Broader Impact}
This paper presents work that aims to advance the fields of language models (LMs) and graph machine learning (GML). Thus, the broader societal impact of this research aligns with prior work in both LMs and GML.

While our proposed approach does not target any specific high-stakes application domain, it may be adopted in settings such as recommendation systems and social network analysis. As such, downstream uses could inherit ethical considerations, including privacy, fairness, and potential misuse. Mitigating such issues is an open challenge shared by the broader community, and we encourage future work to evaluate fairness, robustness, and interpretability as these models are adopted for critical tasks.

\section*{Acknowledgements}
\thanks{This work is supported by NSF (2134079),
and IBM-Illinois Discovery Accelerator Institute. 
The content of the information in this document does not necessarily reflect the position or the policy of the Government, and no official endorsement should be inferred.  The U.S. Government is authorized to reproduce and distribute reprints for Government purposes notwithstanding any copyright notation here on.
}

\bibliography{ref}

\newpage
\appendix
\input{Files/07Appendix}

\end{document}

%% file: Files/00Abstract.tex
\begin{abstract}
Language Models (LMs) are increasingly challenging the dominance of domain-specific models, such as Graph Neural Networks (GNNs) and Graph Transformers (GTs), in graph learning tasks. Following this trend, we propose a novel approach that empowers off-the-shelf LMs to achieve performance comparable to state-of-the-art (SOTA) GNNs on node classification tasks, without any architectural modification. By preserving the LM's original architecture, our approach retains a key benefit of LM instruction tuning: the ability to jointly train on diverse datasets, fostering greater flexibility and efficiency. To achieve this, we introduce two key augmentation strategies: (1) enriching LMs' input using topological and semantic retrieval methods, providing richer contextual information, and (2) guiding the LMs' classification process through a lightweight GNN classifier that effectively prunes class candidates. Experiments on real-world datasets show that backbone Flan-T5 LMs equipped with these augmentation strategies outperform SOTA text-output node classifiers and are comparable to top-performing vector-output node classifiers. By bridging the gap between specialized node classifiers and general LMs, this work paves the way for more versatile and widely applicable graph learning models. We will open-source the code upon publication.

\end{abstract}

%% file: Files/01Intro.tex
\section{Introduction}

There is a growing trend of utilizing Language Models (LMs) for machine learning tasks across diverse domains. This approach has shown tremendous promise in areas such as vision~\citep{DBLP:conf/cvpr/Desai021}, audio~\citep{DBLP:conf/ismir/MittalEHS21}, and multimodal learning~\citep{DBLP:conf/nips/AlayracDLMBHLMM22}. In graph learning, recent efforts have begun to explore the capabilities of LMs in understanding and processing graph structures~\cite{DBLP:conf/acl/0006ZJFJBH025,DBLP:conf/icml/NingFWXH25}.
\citep{DBLP:conf/nips/WangFHTHT23,fu2025bring} showed that LMs can detect node connectivity and identify cycles, while \cite{DBLP:conf/iclr/FatemiHP24} explored LMs' ability to evaluate graph scale and identify connected components. Furthermore, InstructGLM~\citep{DBLP:journals/corr/abs-2308-07134} and LLaGA~\cite{DBLP:conf/icml/Chen0JSW24} achieved state-of-the-art (SOTA) performance in text-output node classifiers on Text-Attributed Graphs (TAG)~\citep{DBLP:conf/www/Zhang0YL24}, whose nodes have textual features.

However, both InstructGLM and LLaGA suffer from a fundamental limitation that \emph{compromises the generality of the backbone LM}. Specifically, InstructGLM expands the LM's vocabulary by creating \emph{a unique token for each node}, whose token embeddings are topology-aware node embeddings. It comes at the cost of incompatibility with two important use cases: (1) multi-task learning on diverse datasets, a common strategy for training Foundational Models~\citep{DBLP:conf/iclr/WeiBZGYLDDL22, DBLP:journals/jmlr/ChungHLZTFL00BW24}, and (2) certain personalized LM fine-tuning services~\citep{DBLP:journals/corr/abs-2401-05459} that restrict access to the backbone model architecture/code\footnote{\url{https://platform.openai.com/docs/guides/fine-tuning}}. LLaGA uses a shared text encoder and a projector to overcome the first limitation but still bears inflexibility when \emph{deploying different LMs} and cannot be applied to LMs without code/architecture access. The above discussion raises a crucial question: {\em How can off-the-shelf, \textbf{text-to-text} instruction-tuned LMs~\citep{DBLP:journals/jmlr/RaffelSRLNMZLL20} achieve competitive performance in node classification tasks without architectural modifications?}

In stark contrast to \citep{DBLP:journals/corr/abs-2309-16595}, which suggests that LMs may only interpret graph structures in prompts as contextual paragraphs, our work presents a more optimistic outlook. We aim to overcome this inherent limitation by augmenting the LMs' input while preserving their original architecture. Our proposed model, \model (\underline{Au}mented \underline{G}raph \underline{L}anguage \underline{M}odel), leverages two key strategies to enhance the LM's ability to process graph:

\begin{itemize}[noitemsep, topsep=0pt]
    \item \textbf{Relevant Node Retrieval}: In contrast to InstructGLM, which relies on multi-hop ego networks akin to message-passing GNNs for structure-aware contextualization,  \model draws inspiration from Graph Transformers (GTs)~\citep{DBLP:journals/corr/abs-2202-08455,DBLP:conf/sigir/Chen0YL00T24,DBLP:conf/iclr/FuHXFZSWMHL24,DBLP:conf/cikm/0007P0CYDT25} and Retrieval-Augmented Generation (RAG)~\citep{DBLP:conf/nips/LewisPPPKGKLYR020, DBLP:conf/icml/GuuLTPC20}. This enables the LM to access long-range structural and semantic information about the target node. We propose two complementary approaches to achieve this: (1) topological retrieval, and (2) prototypical semantic retrieval.  

    \item \textbf{Candidate Label Pruning}: To improve LMs' understanding of graph data while maintaining their text-to-text architecture, we convey the guidance from a specialist model, a pretrained lightweight GNN, to the input of LMs via narrowing down the candidate labels. This allows LMs to focus on discerning between closely related candidates, ultimately enhancing the performance.
\end{itemize}
We extensively evaluate our approach on four real-world TAGs, showing the effectiveness of \modelns. The results indicate that (1) backbone LMs augmented with \model consistently outperform SOTA text-output classifiers while also matching or surpassing the performance of SOTA vector-output classifiers, and (2) \model can be jointly trained on multiple TAGs without performance degradation. These findings represent a crucial step towards bridging the gap between task-specific node classifiers and more general, fine-tuned LMs, highlighting the potential for a unified model excelling in multiple tasks.

%% file: Files/05RelatedWork.tex
\section{Related Work}
Due to the space limitation, the most relevant work is briefly introduced as follows. A more comprehensive literature review is in Section~\ref{sec: additional related work}.

\noindent\textbf{LMs for graphs}. 
Recent studies have explored the ability of LMs to understand graph topology by investigating problems such as graph substructure recall~\citep{DBLP:journals/corr/abs-2402-11821}, connectivity~\citep{DBLP:conf/nips/WangFHTHT23,DBLP:journals/corr/abs-2402-05862}, node/edge counting~\citep{DBLP:journals/corr/abs-2402-05862}, spatial-temporal graphs~\citep{DBLP:conf/kdd/ZhangWZ0Q024}. Building on these findings, several studies have explored tasks on TAGs, including node classification~\citep{DBLP:journals/corr/abs-2308-07134,DBLP:journals/corr/abs-2310-01089,DBLP:journals/corr/abs-2402-03720,DBLP:journals/corr/abs-2310-18152,DBLP:conf/acl/0002WCDZX025,DBLP:journals/corr/abs-2311-14324, DBLP:conf/iclr/WangHZFYCHWYL25}, link prediction~\citep{DBLP:journals/corr/abs-2305-14321, DBLP:journals/corr/abs-2403-04780,DBLP:conf/iclr/0057FKLT0Z24,DBLP:conf/www/0011H0C24, DBLP:conf/icml/TieuF0MH25}, transfer/zero-shot learning~\citep{DBLP:conf/sigir/Tang00SSCY024,DBLP:conf/naacl/SunMFMT25,DBLP:conf/cikm/Pan00H024,DBLP:conf/kdd/HeSHH25,DBLP:journals/corr/abs-2408-10700,DBLP:conf/kdd/0001WLY024}, and graph reasoning~\citep{DBLP:conf/acl/JinXZRZL0TWM024}. GIANT~\citep{DBLP:conf/iclr/ChienCHYZMD22} and GLEM~\citep{DBLP:conf/iclr/0002QLYL00023} encode textual features in a graph-aware way. TAPE~\citep{DBLP:conf/iclr/HeB0PLH24} uses LMs to augment the textual features.


\noindent\textbf{Retrieval-augmented generation (RAG)}
~\citep{DBLP:conf/nips/LewisPPPKGKLYR020, DBLP:conf/emnlp/KarpukhinOMLWEC20} enhances LMs by querying external knowledge~\citep{DBLP:conf/nips/HashimotoGOL18}. This technique retrieves relevant documents from a large corpus and conditioning the LM on the retrieved documents. Building on this, REALM~\cite{DBLP:conf/icml/GuuLTPC20} pretrains the retriever and generator end-to-end. Subsequently, RETRO~\cite{DBLP:conf/icml/BorgeaudMHCRM0L22} scales RAG-enhanced models to large datasets. A crucial component of RAG's success is the objective function proposed by~\cite{DBLP:journals/corr/abs-2301-12652}, which enables the training of retriever even the LMs is a block box. HyDE~\cite{DBLP:conf/iclr/0002IWXJ000023} uses hypothetical document generation to improve retrieval in RAG systems. Furthermore, RAG has extended to multimodal settings~\cite{DBLP:conf/icml/YasunagaAS0LLLZ23}. Recently, GraphRAG~\citep{edge2024local} has garnered significant attention; it constructs a Knowledge Graph and then generates responses based on the summaries of communities derived from KG. These advancements have significantly improved generated text's accuracy and contextual relevance.




%% file: Files/02Prelim.tex
\section{Preliminaries}
We use the following notation conventions: bold lower-case letters (e.g,. $\mathbf x$) denote column vectors, bold upper-case letters (e.g., $\mathbf X$) denote matrices, and calligraphic upper-case letters (e.g., $\mathcal X$) denote sets. We use $[\cdot]$ and $[\cdot,\cdot]$ to index vectors and matrices, respectively.

We study the node classification problem~\cite{DBLP:journals/kais/XuDWLT24,DBLP:conf/icml/0002QZYZ0ZWHT24,DBLP:conf/www/XuDT22} on TAGs where each node is associated with textual attributes. A TAG with $n$ nodes is represented as $\mathcal G = (\mathcal V, \mathcal E, \mathcal T)$, where $\mathcal V=\{v_i\}_{i=1}^n$ denotes a set of nodes, and $\mathcal E=\{e_{ij}\}_{i,j=1}^n$ is a set of edges where $e_{ij}=1$ indicates that nodes $v_i$ and $v_j$ are connected; otherwise, $e_{ij}=0$. $\mathcal T=\{t_i\}_{i=1}^n$ indicates the set of node textual attributes. The edges can also be represented by an adjacency matrix $\mathbf A\in\{0,1\}^{n\times n}$, where $\mathbf A[i,j] = 1$ if and only if $e_{ij}=1$. The training and test node labels are denoted by $\mathcal Y = \mathcal Y_{\mathrm{train}}\cup\mathcal Y_{\mathrm{test}}=\{y_i\}_{i=1}^n$, where each label $y_i$ belongs to one of the $C$ classes, i.e., $y_i\in\{1,\dots,C\}, \forall i$. In the semi-supervised setting studied in this paper, the graph structure and training labels $\mathcal V, \mathcal E, \mathcal T, \mathcal Y_{\mathrm{train}}$ are accessible, and the goal is to predict the labels of test nodes $\mathcal Y_{\mathrm{test}}$.

\begin{figure*}[t!]
\centering
\begin{subfigure}{0.46\textwidth}
  \centering
  \includegraphics[width=\textwidth]{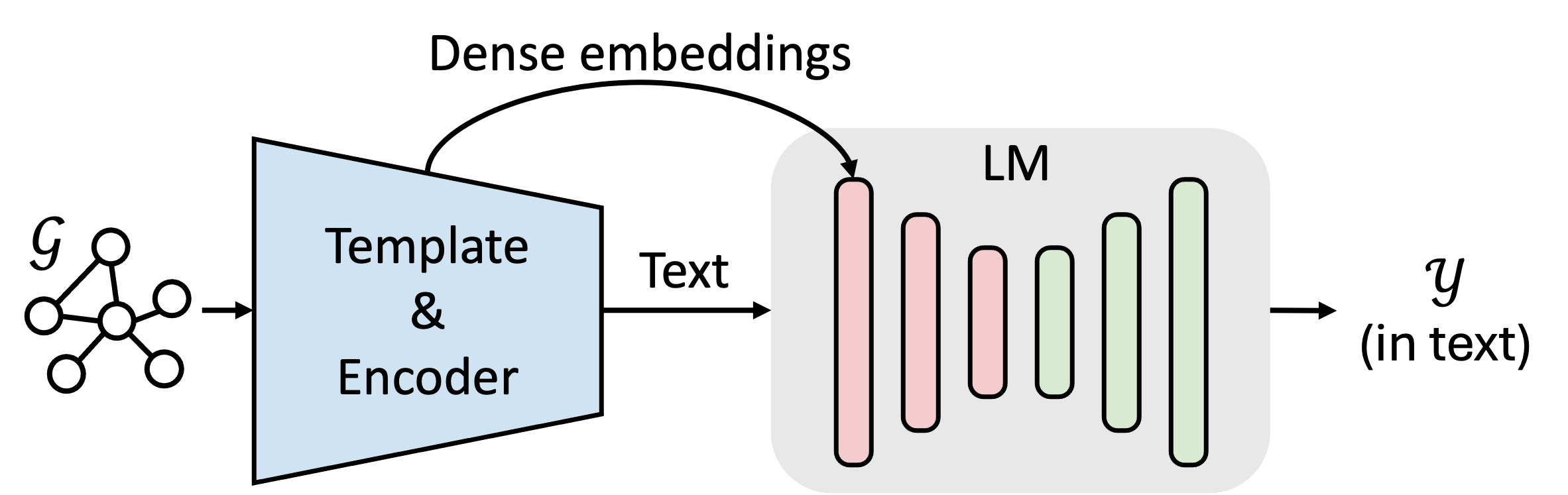}
  \caption{InstructGLM and LLaGA.}
  \label{fig: InstructGLM pipeline}
\end{subfigure}
\hspace{3mm}
\begin{subfigure}{0.46\textwidth}
  \centering
  \includegraphics[width=\textwidth]{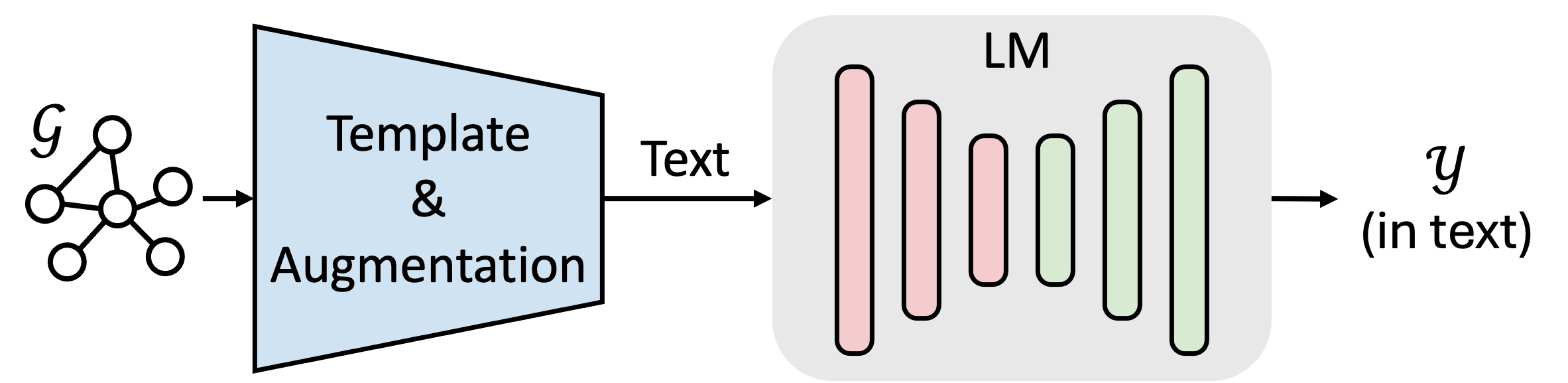}
  \caption{\model (ours).}
  \label{fig: our pipeline}
\end{subfigure}
\vspace{-2mm}
\caption{Comparison of pipelines between the existing LM-based node classifiers and our approach, \modelns. Unlike InstructGLM and LLaGA, which explicitly encodes graph information into token embeddings as a form of soft prompting, \model maintains the original text-to-text framework of the off-the-shelf LM, offering greater generality and flexibility.}
\label{fig: comparison between SOTA and ours}
\end{figure*}

\noindent\textbf{Personalized PageRank (PPR)}
\label{sec: ppr}
~\citep{page1999pagerank,DBLP:conf/www/JehW03} ranks all the nodes according to their relevance to a given query node. Specifically, given the adjacency matrix $\mathbf A$, the PPR scores $\mathbf r_i \in \mathbb R^{n}$ for all nodes concerning the query node $v_i$ are computed iteratively as:
\begin{align}
    \mathbf r_i\leftarrow (1-\alpha) \tilde{\mathbf A} \mathbf r_i + \alpha \mathbf q_i
    \label{eq: ppr computation}
\end{align}
where $\alpha\in(0,1)$ is the teleport probability, $\mathbf q_i\in \{0,1\}^{n}$ is a one-hot vector whose $i$-th entry is $1$, $\tilde{\mathbf A}=\mathbf A\mathbf D^{-1}$ is the normalized adjacency matrix, and $\mathbf D$ is the degree matrix. Once $\mathbf r_i$ converges, the top-$K$ relevant nodes concerning the query node $v_i$ can be identified as follows: 
\begin{align}
    \mathtt{PPR}(v_i, K) = \{v_j: \mathbf r_i[j]\in \mathrm{topK} (\mathbf r_i)\}
    \label{eq: ppr selection}
\end{align}

\noindent\textbf{Language models (LMs).}
We employ autoregressive LMs that predict the next token $z_i$ based on the input sequence $t$ and the context of previously generated tokens $z_{1:i-1}$. The probability of generating a sequence $z$ given the input $t$ is:
\begin{align}
    p_{\mathrm{LM}}(z|t) = \prod_{i=1}^{|z|} p_{\mathrm{LM}}(z_i|t,z_{1:i-1})
    \label{eq: LM generation prob}
\end{align}
\textbf{Retrieval-augmented generation (RAG)}
~\citep{DBLP:conf/nips/LewisPPPKGKLYR020, DBLP:conf/icml/GuuLTPC20} first retrieves a query $t$-relevant text $d^*$ from an external corpus $\mathcal D$ via a similarity function $s_{\phi}$:
\begin{align}
    d^* = \mathop{\arg\max}_{d\in\mathcal D} s_{\phi}(d, t)
    \label{eq: RAG retrieve}
\end{align}
$s_{\phi}$ is typically implemented as a dual-encoder architecture~\citep{DBLP:journals/ijprai/BromleyBBGLMSS93}:
\begin{align}
    s_\phi (d,t) = \left\langle\mathtt{Encoder}_{\phi}(d), \mathtt{Encoder}_{\phi}(t)\right\rangle
    \label{eq: dual encoder}
\end{align}
Once $d^*$ is retrieved, it is fed into the LM together with the query $t$: $p_{\mathrm{LM}}(z|d^*, t)$ for generation.

%% file: Files/03Method.tex
\section{Method}

\begin{figure*}[t!]
\centering
\begin{subfigure}{0.36\textwidth}
  \centering
  \includegraphics[width=\textwidth]{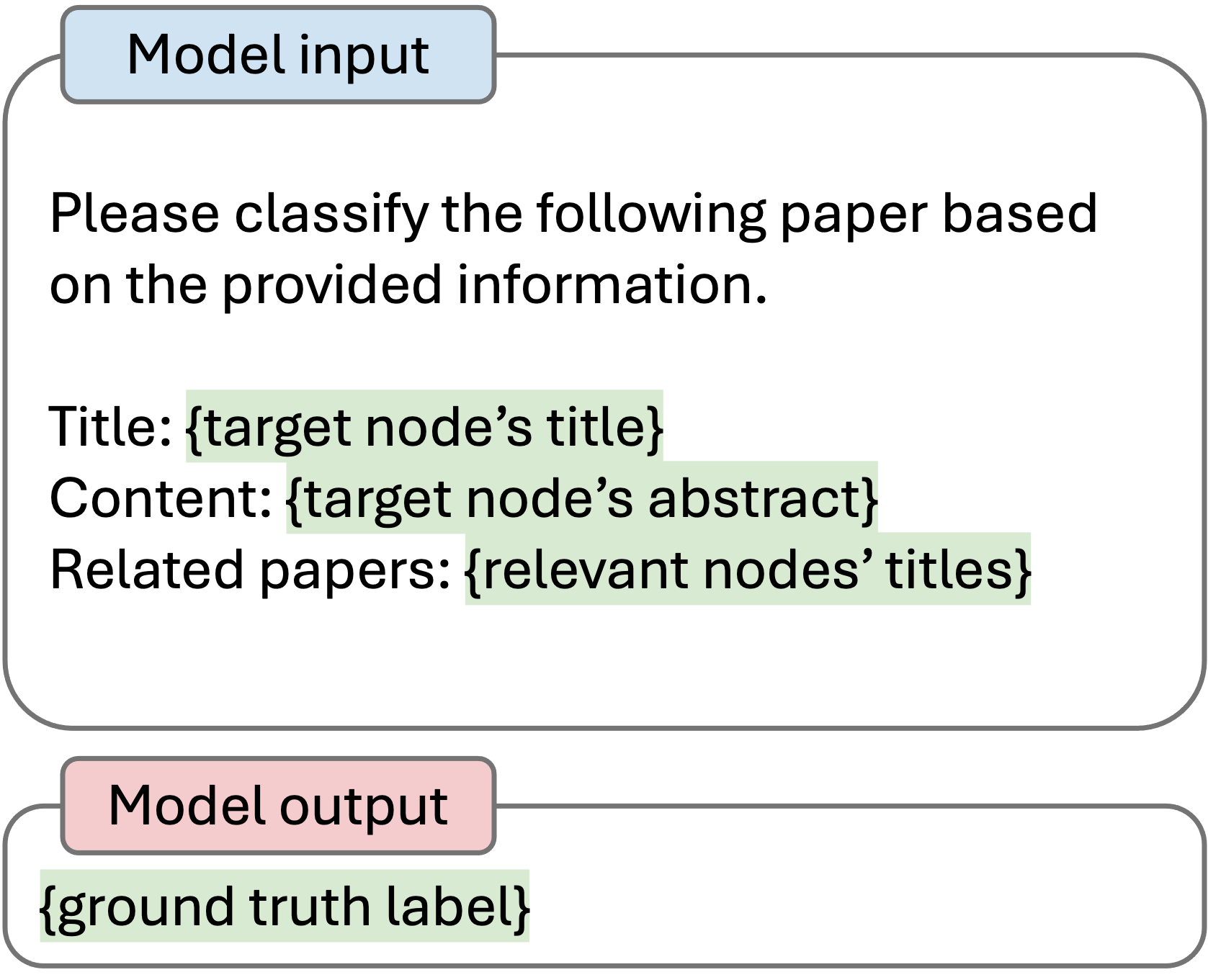}
  \caption{A typical graph-to-text template.}
  \label{fig: others template}
\end{subfigure}
\hspace{5mm}
\begin{subfigure}{0.36\textwidth}
  \centering
  \includegraphics[width=\textwidth]{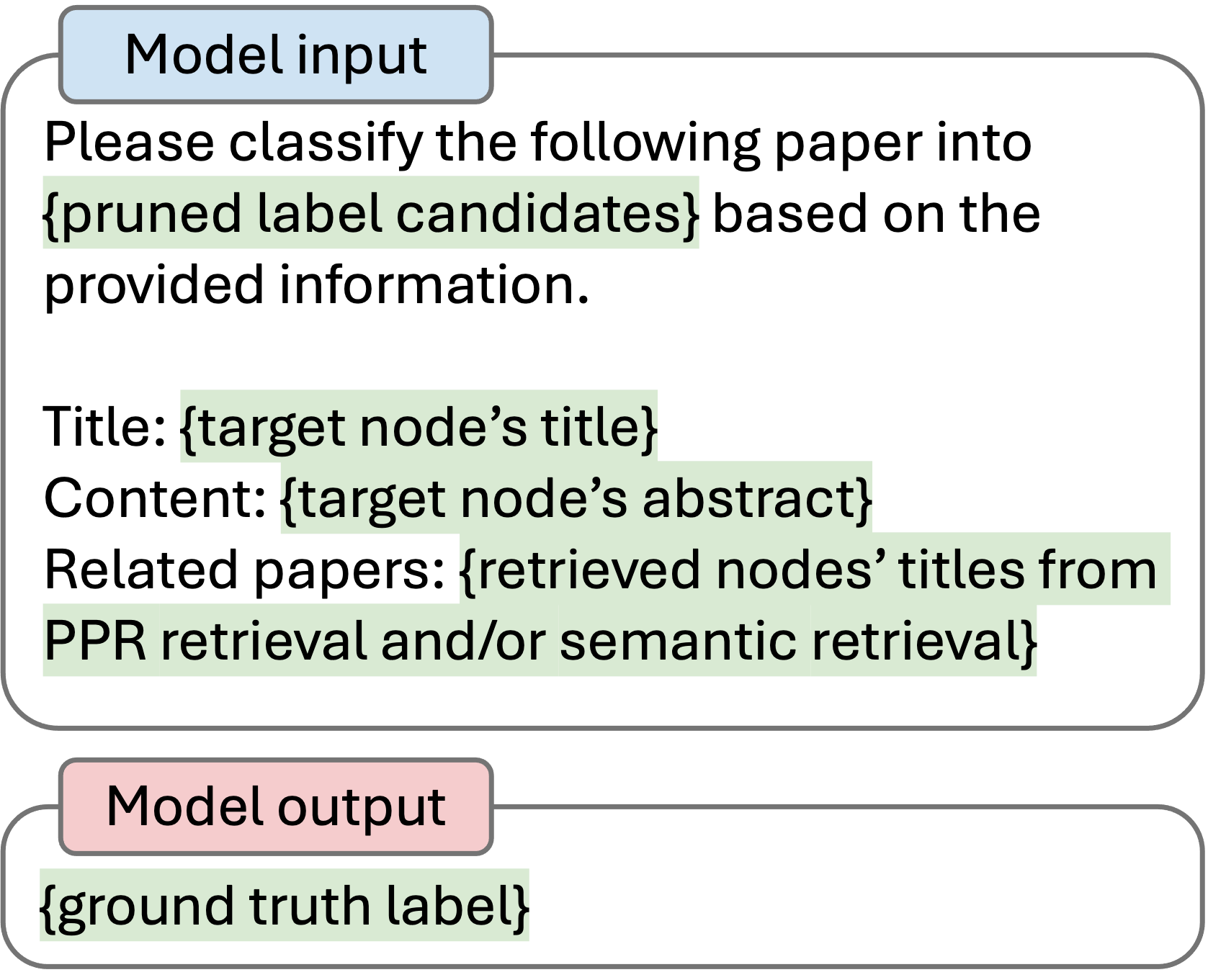}
  \caption{Our template with augmented text.}
  \label{fig: our template}
\end{subfigure}
\vspace{-2mm}
\caption{Comparison of a typical graph-to-text template (a) and our template with augmented text features (b).}
\label{fig: template comparison}
\end{figure*}

\begin{figure*}[t!]
\centering
\includegraphics[width=.9\textwidth]{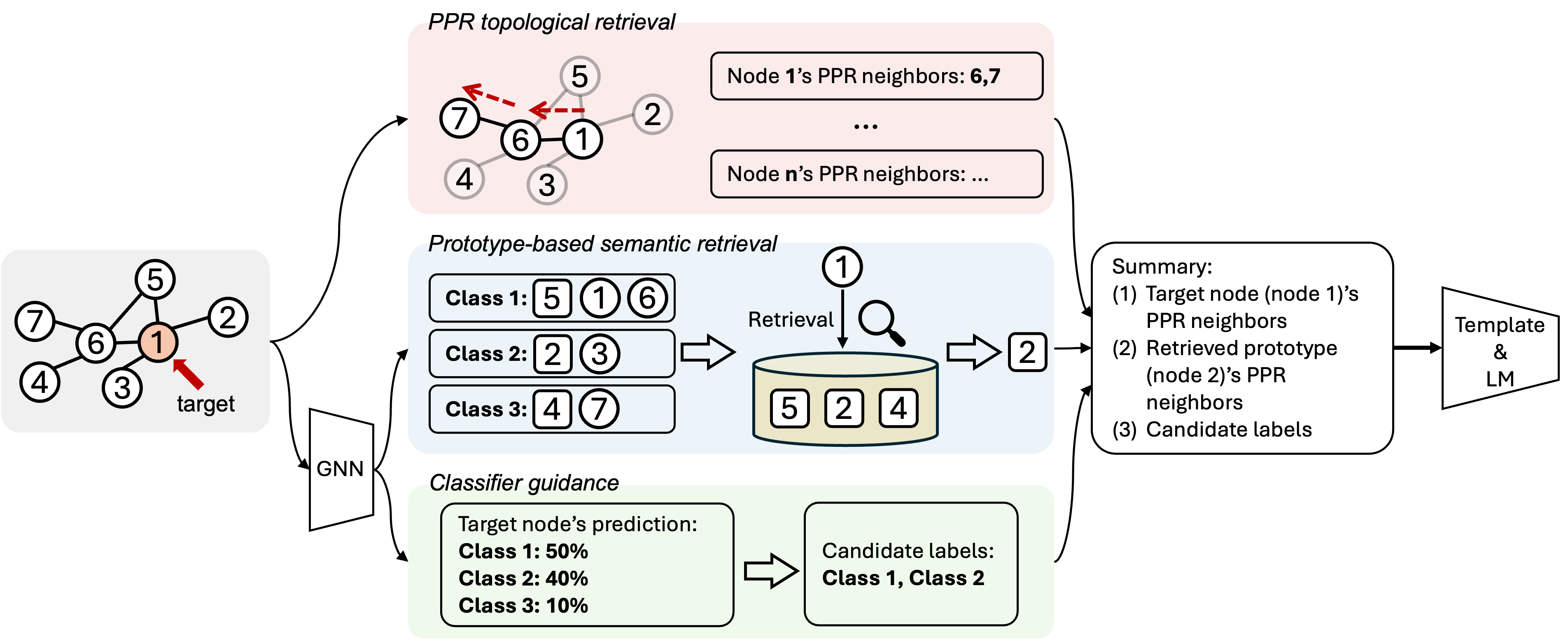}
\vspace{-2mm}
\caption{A detailed pipeline of \modelns. In the semantic retrieval module, rectangles denote the class prototypes.}
\label{fig: detailed pipeline}
\end{figure*}

We explore the application of LMs to node classification by reformulating it as a text-to-text task~\cite{DBLP:journals/jmlr/RaffelSRLNMZLL20}. Our method employs carefully designed prompt templates and augmentation techniques to transform graph and ground truth labels into text pairs, enabling LMs to process and be fine-tuned {\em without} modifying their underlying architecture.

As shown in Figure~\ref{fig: comparison between SOTA and ours}, \model fundamentally differs from InstructGLM~\citep{DBLP:journals/corr/abs-2308-07134} and LLaGA~\cite{DBLP:conf/icml/Chen0JSW24}, the current SOTA LM node classifiers. While all three methods utilize prompt templates to transform input graphs into text, InstructGLM and LLaGA explicitly \textbf{encode node features into the LM's token embeddings} which can be categorized as \textbf{soft prompting}~\citep{DBLP:conf/emnlp/LesterAC21}. In contrast, our approach provides a \textbf{data augmentation}-based framework without modifying LM's text-to-text architecture, enabling our model to retain the versatility of the original LM. The following section first details the augmentation techniques developed by this paper and then introduces the templates to incorporate all the augmented textual features. Figure~\ref{fig: detailed pipeline} provides an overview of all modules developed in this paper.

\subsection{Retrieval-based aggregation}
\label{sec: retriever}
General LMs are not designed to process graph data directly. To overcome this, a common approach is to employ prompt templates to transform graphs and associated tasks into text that LMs can understand. For instance, for the Cora~\citep{DBLP:journals/aim/SenNBGGE08} literature citation graph, a typical template~\citep{DBLP:journals/corr/abs-2309-16595, DBLP:journals/corr/abs-2308-07134} for node classification, as shown in Figure~\ref{fig: others template}, consists of three main components: (1) a short task description, (2) the target node's textual features, e.g., its title and abstract, and (3) textual features from relevant nodes.


The success of the message-passing GNNs highlights the importance of the \textbf{aggregation} operation, whose typical example is the mean pooling of intermediate node embeddings. A similar spirit is followed for the LM-based classifiers, whose key design is the \textbf{selection of relevant nodes}. Existing works~\citep{DBLP:journals/corr/abs-2309-16595, DBLP:journals/corr/abs-2308-07134} select one/multi-hop neighbors as relevant nodes, but we posit that this approach is suboptimal for two reasons. Firstly, not all immediate or extended neighbors are relevant to the target node, which can introduce noise and degrade model performance. Secondly, incorporating multi-hop neighbors can lead to "neighbor explosion"~\citep{DBLP:conf/nips/HamiltonYL17,DBLP:conf/icml/ChenZS18,DBLP:conf/icml/FeyLWL21}, i.e., an exponentially-growing set of "relevant" nodes, resulting in increased computational costs and even leading to the out-of-memory issue. As a response, two novel solutions, \emph{topological retrieval} and \emph{prototypical semantic retrieval}, are proposed to efficiently identify the most informative nodes for the classification tasks.

\noindent\textbf{Topological retrieval.}  PPR~\citep{page1999pagerank,DBLP:conf/www/JehW03} is leveraged for topological retrieval, which has shown great effectiveness in conjunction with GNNs~\citep{DBLP:conf/iclr/KlicperaBG19,DBLP:journals/corr/abs-2412-16435,DBLP:conf/icdm/XuDW0T22,DBLP:conf/icml/XuYW0ZA0T24,DBLP:conf/kdd/XuCZWPYT23,DBLP:conf/wsdm/FuXTH23}. The success of PPR suggests that its retrieved neighbors may provide more informative context than generic one/multi-hop neighbors. Specifically, for a target node $v_i$,  we select its top-$K$ neighbors $\mathtt{PPR}(v_i, K)$ based on their PPR scores (Eqs.~(\ref{eq: ppr computation}) and~(\ref{eq: ppr selection}), Section~\ref{sec: ppr}). Then, the text features from the PPR neighbors are concatenated as the PPR-retrieved text $t_{\mathrm{PPR}} = \mathop{\oplus}_{j;v_j\in \mathtt{PPR}(v_i, K)} t_j$, where $\oplus$ denotes text concatenation.

It is worth noting that the classic PPR algorithm is computationally expensive for large graphs due to the matrix multiplication (Eq.~(\ref{eq: ppr computation})). However, efficient approximate solutions such as ApproximatePR~\citep{DBLP:conf/focs/AndersenCL06}, can be applied to mitigate this issue. Nevertheless, PPR is a topology-based heuristic that inherently cannot leverage textual features or supervision from downstream LMs. To enhance our framework's semantic awareness, we propose a complementary strategy called prototypical semantic retrieval as follows.

\noindent\textbf{Prototypical semantic retrieval.} Our semantic retrieval module draws inspiration from two popular techniques: (1) RAG~\citep{DBLP:conf/nips/LewisPPPKGKLYR020, DBLP:conf/icml/GuuLTPC20}, which retrieves external corpora, and (2) Graph Transformers~\citep{DBLP:journals/corr/abs-2202-08455}, which aggregate messages from distant nodes via inner product-based attention weights. For node classification, we treat \textbf{the textual features of all nodes except the target node} as a surrogate "external corpus." However, unlike typical question-answering tasks, retrieving textual features from \textbf{a single node is often insufficient} for accurate node classification. To address this, we enhance the semantic retrieval by retrieving prototypes, which capture the essence of each class~\citep{snell2017prototypical}.

Prototypes~\citep{biehl2016prototype} are defined as representative examples in classification problems. To obtain prototypes, a lightweight GNN $\psi$ is pretrained and generates a prediction vector for each node: $\tilde{\mathbf{y}}_i = \mathtt{GNN}_\psi(v_i, \mathcal G)\in\mathbb R^C, \forall v_i$. The prediction confidence for each node $v_i$ is defined as: $\mathtt{Conf}(v_i) = \max_j \tilde{\mathbf y}_i[j]$.

The predicted class-$c$ examples are $\tilde{\mathcal Y}_c = \{v_i: \mathop{\arg\max}_j \tilde{\mathbf y}_i[j] = c\}$ and their confidence is $\mathtt{Conf}_c=\{\mathtt{Conf}(v_i): v_i\in \tilde{\mathcal Y}_c\}$. For each class $c$, the top-$N$ confident examples are selected as prototypes:
\begin{align}
    \mathcal P_c = \left\{ v_i: v_i\in \tilde{\mathcal Y}_c \land \mathtt{Conf}(v_i)\in \mathrm{topN}\left(\mathtt{Conf}_c\right) \right\}
    \label{eq: prototype selection}
\end{align}
For all the classes, there are $N\times C$ prototypes: $\mathcal P = \bigcup_{c\in \{1,\dots, C\}}\mathcal P_c$. To ensure every document in the corpus $\mathcal D$ includes text features from \textbf{multiple nodes}, $\mathcal D$ is constructed by concatenating the text of \textbf{each prototype's PPR neighbors}:
\begin{align}
    \mathcal D = \left\{\mathop{\oplus}_{j;v_j\in \mathtt{PPR}(v_i, K)} t_j: v_i \in \mathcal P\right\}
    \label{eq: prototype text}
\end{align}
Next, for each target node with its associated text features $t_{\mathrm{target}}$, we compute the prototypically retrieved text using Eq.~(\ref{eq: RAG retrieve}): $t_{\mathrm{proto}} = \mathop{\arg\max}_{d\in\mathcal D} s_{\phi}(d, t_{\mathrm{target}})$. In our experiments, we may use $t_{\mathrm{PPR}}$ (from topological retrieval), or $t_{\mathrm{proto}}$, or both by concatenation $t_{\mathrm{PPR}}\oplus t_{\mathrm{proto}}$. For simplicity, we denote the final retrieved text as $t_{\mathrm{retri}}$.

\subsection{Classifier guidance}
\label{sec: classifier guidance}
Recent studies~\citep{DBLP:journals/corr/abs-2309-16595,DBLP:conf/iclr/FatemiHP24,DBLP:conf/kdd/00010TL24} highlighted mainstream LMs' limited understanding of graph topology. While InstructGLM~\citep{DBLP:journals/corr/abs-2308-07134} and LLaGA~\citep{DBLP:conf/icml/Chen0JSW24} address this limitation by incorporating topology-aware node embeddings (e.g., from a pretrained or parameter-free GNN) into the LM's token embeddings, this approach necessitates \textbf{modifications to the LM's architecture}. We propose an alternative method that conveys guidance from a pretrained GNN into the \textbf{input text of LMs}, thereby preserving the LM's original architecture. Concretely, such guidance is to \textbf{prune the classification candidates}.

We repurpose the pretrained $\mathtt{GNN}_\psi$ from the prototypical semantic retrieval module. For each node $v_i$, we identify and save the top-$I$ predicted labels:
\begin{align}
    \mathcal L_i = \left\{j: \tilde{\mathbf y}_i[j]\in\mathrm{topI}\left( \tilde{\mathbf y}_i\right)\right\}\in \{1,\dots,C\}^I
    \label{eq: pruned label candidates}
\end{align}
where $I < C$. For datasets in the experiments, the $\mathtt{IndexToLabel}$ maps are available, which map numerical labels to their corresponding text. The pruned label candidates for node $v_i$ can be presented as concatenated text: $t_{\mathrm{candidates}}=\oplus_{i\in \mathcal L_i}\mathtt{IndexToLabel}(i)$. The integration of pruned candidates into the template is detailed in Section~\ref{sec: overall template}.

By focusing on a more relevant set of candidate labels, valuable topology-aware inductive bias from the GNN is incorporated into the LM's input, thereby enhancing its node classification performance without altering its architecture.

\subsection{Overall template}
\label{sec: overall template}
Our augmented training samples include three key elements: (1) the target node's text $t_{\mathrm{target}}$, (2) the retrieved nodes' text $t_{\mathrm{retri}}$, and (3) the pruned label candidates $t_{\mathrm{candidates}}$. We collectively denote these elements as $t_{\mathrm{in}}=(t_{\mathrm{target}}, t_{\mathrm{retri}}, t_{\mathrm{candidates}})$. LM's prediction probability for the target sequence $y_{\mathrm{target}}$ is based on Eq.~\eqref{eq: LM generation prob} whose input text $t$ is $t_{\mathrm{in}}$.
Figure~\ref{fig: our template} presents an exemplar template for the Cora dataset, showcasing the integration of $t_{\mathrm{target}}$,  $t_{\mathrm{retri}}$, and $t_{\mathrm{candidates}}$. The selection of the backbone LM will be detailed in Section~\ref{sec: exp}.  Appendix~\ref{sec: templates} contains a full list of templates. Note that we exclude the "abstracts" of the retrieved nodes to prevent exceeding the maximum input length of most LMs. We utilize only this template's "model input" part for evaluation.



\subsection{Training}
\label{sec: training}

Our framework includes three parameterized modules that require training or fine-tuning: (1) GNNs for generating prototypes and candidate label pruning, as described in Sections~\ref{sec: retriever} and~\ref{sec: classifier guidance}, (2) the encoder $\phi$ from the semantic retriever, defined in Eq.~\ref{eq: dual encoder}, and (3) the backbone LM, utilized in Eq.~\ref{eq: LM generation prob}. The GNNs from Sections~\ref{sec: retriever} and~\ref{sec: classifier guidance} can be shared, and their training is independent of the other modules, which are supervised by ground truth labels. This process is detailed in Appendix~\ref{sec: intro of GNN}.

One of the standard LM's losses, the average token-wise negative log-likelihood (NLL), is used. For a target node, the loss is:
\begin{align}
    \mathcal L_{\mathrm{NLL}} \left(p_{\mathrm{LM}}(y_{\mathrm{target}}|t_{\mathrm{in}}), y_{\mathrm{target}}\right)
    \label{eq: NLL for finetuning backbone LM}
\end{align}
To train the semantic retriever, we employ a distribution-matching loss. For a given target node's text $t_{\mathrm{target}}$, its retrieval probability for a prototype text $t\in \mathcal D$ is:
\begin{align}
    p_{\phi}(t|t_{\mathrm{target}}) = \frac{e^{s_\phi (t,t_{\mathrm{target}})}}{\sum_{t'\in\mathcal D}e^{s_\phi (t',t_{\mathrm{target}})}}
    \label{eq: retrieval distribution}
\end{align}
Next, an LM-supervised empirical distribution is:
\begin{align}
    \tilde{p}_{\mathrm{LM}}(t|t_{\mathrm{target}}, y_{\mathrm{target}}) = \frac{e^{p_{\mathrm{LM}}(y_{\mathrm{target}}|t_{\mathrm{in}})}}{\sum_{t'\in\mathcal D}e^{p_{\mathrm{LM}}\left(y_{\mathrm{target}}|t_{\mathrm{in}}'\right)}}
    \label{eq: LM supervision distribution}
\end{align}
where $t_{\mathrm{in}}=(t_{\mathrm{target}}, t, t_{\mathrm{candidates}})$ and $t_{\mathrm{in}}'=(t_{\mathrm{target}}, t', t_{\mathrm{candidates}})$. This distribution represents the \textbf{normalized importance of each prototype text} $t\in\mathcal D$ based on the \textbf{LM's likelihood of generating the ground truth text} $y_{\mathrm{target}}$. We use $\tilde{p}_{\mathrm{LM}}$ to distinguish this distribution from the generation probability in Eq.~(\ref{eq: LM generation prob}). The distribution matching loss is the Kullback-Leibler (KL) divergence between the retrieval and the LM-supervised distributions:
\begin{align}
    \mathrm{KL}\left(\mathtt{sg}\left(\tilde{p}_{\mathrm{LM}}\left(\cdot|t_{\mathrm{target}},y_{\mathrm{target}}\right)\right)\|p_{\phi}(\cdot|t_{\mathrm{target}})\right)
    \label{eq: KL loss}
\end{align}
This loss aims to align the retrieval probability of prototype text $t\in\mathcal D$ with its importance in facilitating the LM's generation of the label text $y_{\mathrm{target}}$ for the target node. The stop gradient operator $\mathtt{sg}$ ensures that the loss Eq.~\eqref{eq: KL loss} only updates the semantic retriever $\phi$ but keeps the LM's parameters $\theta$ frozen. This objective has been used by previous works~\citep{DBLP:journals/corr/abs-2301-12652,
izacard2023atlas} without a thorough analysis. We provide an in-depth examination of its implications in Appendix~\ref{sec: interpret distribution matching loss}.

Notably, Eq.~(\ref{eq: LM supervision distribution}) requires $|\mathcal D|$  inferences of the LM due to the denominator. However, the LM is fine-tuned via the NLL loss, Eq.~(\ref{eq: NLL for finetuning backbone LM}), only for the most relevant prototype, $\mathop{\arg\max}_{d\in\mathcal D} s_{\phi}(d, t_{\mathrm{target}})$. Consequently, each update step involves $|\mathcal D|$ forward passes but only one backpropagation. To reduce the computational overhead associated with $|\mathcal D|$ inferences, we can use a sampling strategy: selecting the top-$M$ samples to form a retrieval minibatch $\mathcal D_M = \{t: t\in \text{topM}_{t'\in\mathcal D} s_\phi (t',t_{\mathrm{target}})\}$. By replacing $\mathcal D$ with $\mathcal D_M$ in Eqs.~(\ref{eq: retrieval distribution}) and~(\ref{eq: LM supervision distribution}), the retrieval and the LM-supervised distributions can be computed "in-batch", reducing the inference times from $|\mathcal D|$ to $M$.

Algorithm~\ref{alg: training} (Appendix~\ref{sec: algorithm}) outlines a step-by-step process for fine-tuning \modelns, processing one training node per step. This procedure can be readily extended to mini-batch settings.

\subsection{Model complexity}
\model consists of three parameterized modules: (1) a GNN $\psi$, (2) the semantic retriever $\phi$, and (3) the backbone LM $\theta$. Notably, $\psi$ and $\phi$ are lightweight, whose number of parameters is only $1/30$ to $1/3$ of the number of LM $\theta$ parameters. Compared to the SOTA InstructGLM~\cite{DBLP:journals/corr/abs-2308-07134}, \model has an additional module $\phi$, resulting in a slightly increased number of parameters. For training, the GNN $\psi$ can be pretrained, and the PPR scores can be precomputed. The training of $\theta$ relies on the retrieved text from $\phi$, while the training of $\phi$ requires $\tilde{p}_{\mathrm{LM}}(\cdot|t_{\mathrm{target}}, y_{\mathrm{target}})$, which is obtained through forward inference of $\theta$. Importantly, computational graphs (used for gradient computation) of $\theta$ and $\phi$ are \textbf{independent}. When training $\phi$, the stop gradient operator $\mathtt{sg}$ ensures  $\theta$ has no gradient. As a result, the cost of backpropagation is close to the sum of the cost for updating the LM $\theta$ and the semantic encoder $\phi$, separately.



\subsection{Discussion on the flexibility of \modelns}
We noticed that LLaGA~\cite{DBLP:conf/icml/Chen0JSW24} improves the generality of soft prompting-based solutions by a shared text encoder and a projector. Here, we compare the flexibility and generality of LLaGA and our proposed \model to illustrate our unique contribution better.
\begin{itemize}[noitemsep, topsep=0pt]
    \item LlaGA \textbf{can achieve $0$-shot transfer learning}, e.g., training on the Cora dataset and testing on the Amazon dataset. However, LLaGA \textbf{cannot switch the LM seamlessly} because it requires modifying the LM's code and architecture, e.g., including its graph encoder's output into the LM's token embeddings. The application of LLaGA is also limited if there is no access to the LM's architecture or code.
    \item Our \model \textbf{cannot achieve $0$-shot transfer learning} because we need a trained GNN to provide reliable label candidates for text input augmentation. However, thanks to such a data augmentation paradigm, \model \textbf{can switch the LM seamlessly} as long as the LM works in a text-to-text manner.
\end{itemize}


%% file: Files/04Exp.tex
\begin{table*}[t]
\centering
\resizebox{0.76\linewidth}{!}{
\begin{tabular}{ll|llll}
\toprule
 & \textbf{Method} & \textbf{Cora} & \textbf{Pubmed} & \textbf{ogbn-arxiv} & \textbf{ogbn-products} \\
\midrule
\multirow{11}{*}{\rotatebox{90}{Vector-output}} 
 & GCN                 & 87.78\scriptsize$\pm$0.96 & 88.90\scriptsize$\pm$0.32 & 73.60\scriptsize$\pm$0.18 & 75.64\scriptsize$\pm$0.21 \\
 & GraphSAGE           & 86.51\scriptsize$\pm$2.36 & 89.08\scriptsize$\pm$0.28 & 73.88\scriptsize$\pm$0.33 & 76.04\scriptsize$\pm$0.25 \\
 & GLEM + RevGAT       & 88.56\scriptsize$\pm$0.60 & 94.71\scriptsize$\pm$0.20 & 76.97\scriptsize$\pm$0.19 & -- \\
 & GIANT + RevGAT      & 83.53\scriptsize$\pm$0.38 & 85.02\scriptsize$\pm$0.48 & 75.90\scriptsize$\pm$0.19 & 71.89\scriptsize$\pm$0.30 \\
 & GIANT + GCN         & 84.23\scriptsize$\pm$0.53 & 84.19\scriptsize$\pm$0.50 & 73.29\scriptsize$\pm$0.10 & 69.77\scriptsize$\pm$0.42 \\
 & TAPE + RevGAT       & \textbf{\color{blue}92.90\scriptsize$\pm$3.07} & \textbf{\color{blue}96.18\scriptsize$\pm$0.53} & 77.50\scriptsize$\pm$0.12 & \textbf{\color{blue}82.34\scriptsize$\pm$0.36} \\
 & ENGINE       & 91.48\scriptsize$\pm$0.32 & -- & 76.02\scriptsize$\pm$0.29 & -- \\
 & SimTeG+RevGAT       & -- & -- & 77.04\scriptsize$\pm$0.13 & -- \\
 & GraphAdapter       & -- & -- & 77.07\scriptsize$\pm$0.15 & -- \\
 & OFA       & 74.76\scriptsize$\pm$1.12 & 78.21\scriptsize$\pm$0.71 & \textbf{\color{blue}77.51\scriptsize$\pm$0.17} & -- \\
 & LLM4GT   & 89.16\scriptsize$\pm$0.96 & 94.72\scriptsize$\pm$0.56 & -- & -- \\
\midrule
\multirow{6}{*}{\rotatebox{90}{Text-output}} 
 & DeBERTa             & 76.06\scriptsize$\pm$3.78 & 94.94\scriptsize$\pm$0.46 & 73.61\scriptsize$\pm$0.04 & 72.97\scriptsize$\pm$0.23 \\
 & GraphPrompter & 82.26 & 94.80 & 75.61 & 79.54 \\
 & InstructGLM         & 90.77\scriptsize$\pm$0.52 & 94.62\scriptsize$\pm$0.13 & 75.70\scriptsize$\pm$0.12 & -- \\
 & LLaGA               & 89.85 & 95.06 & 76.66 & -- \\
 & GraphICL               & 83.58 & 93.18 & 73.68 & 81.48 \\
 & \model{} (T5-small) & 91.14\scriptsize$\pm$0.55 & 94.80\scriptsize$\pm$0.15 & 75.39\scriptsize$\pm$0.21 & 81.73\scriptsize$\pm$0.08 \\
 & \model{} (T5-base)  & 91.24\scriptsize$\pm$0.46 & 95.03\scriptsize$\pm$0.35 & \textbf{\color{red}76.80\scriptsize$\pm$0.14} & 81.91\scriptsize$\pm$0.11 \\
 & \model{} (T5-large) & \textbf{\color{red}91.51\scriptsize$\pm$0.26} & \textbf{\color{red}95.16\scriptsize$\pm$0.18} & 76.00\scriptsize$\pm$0.23 & \textbf{\color{red}82.90\scriptsize$\pm$0.10} \\
\bottomrule
\end{tabular}
}
\vspace{-2mm}
\caption{Accuracy (\%) comparison between \model{} and existing SOTA models. The best-performing \textbf{\textcolor{blue}{vector-output}} and \textbf{\textcolor{red}{text-output}} models on each dataset are highlighted in \textbf{\textcolor{blue}{blue}} and \textbf{\textcolor{red}{red}}, respectively.}
\label{tab: comparison with SOTAs}
\end{table*}

\section{Experiments}
\label{sec: exp}

This section introduces the experimental setups, baseline methods, effectiveness studies, ablation studies, and multi-task training. Efficiency and hyperparameter studies are in the Appendix.

\subsection{Setup}
Following \citep{DBLP:conf/iclr/HeB0PLH24, DBLP:journals/corr/abs-2308-07134}, we evaluate our approach on four benchmark datasets: Cora~\citep{DBLP:journals/aim/SenNBGGE08}, Pubmed~\citep{DBLP:journals/aim/SenNBGGE08}, ogbn-arxiv~\citep{DBLP:journals/corr/abs-2005-00687}, and a subset of ogbn-products~\citep{DBLP:journals/corr/abs-2005-00687,DBLP:conf/iclr/HeB0PLH24}. The dataset statistics are in Table~\ref{tab: dataset statistics}.

Our implementation employs two pretrained all-MiniLM-L6-v2 models~\cite{wang2020minilm} as the the semantic retriever $\phi$ (Eq.~(\ref{eq: dual encoder})) and the text encoder for GNN $\psi$ (Eq.~(\ref{eq: encoder of GNN})). We set the PPR teleport probability $\alpha=0.1$. We employ a $3$-layer GraphSAGE~\citep{DBLP:conf/nips/HamiltonYL17} with a hidden dimension of $256$ as $\psi$. Our hyperparameters include $K=5$ PPR neighbors, $N=10$ prototypes, and $M=8$ samples for LM inference. The number of label candidates $I$ is searched from $\{2,3\}$. Flan-T5-small/base/large~\citep{DBLP:journals/corr/abs-2210-11416} are used as the backbone LM $\theta$ with templates detailed in Section~\ref{sec: templates}.


\subsection{Comparison with state-of-the-arts}




This section presents the comparison between \model and SOTA baselines. We categorize models into two groups: (1) vector-output models which output a vector with dimension equal to the number of classes, and (2) text-output models, whose output is text. Specifically, results from GCN~\citep{DBLP:conf/iclr/KipfW17}, GraphSAGE~\cite{DBLP:conf/nips/HamiltonYL17}, GLEM~\citep{DBLP:conf/iclr/0002QLYL00023}+RevGAT~\citep{DBLP:conf/icml/Li0GK21}, InstructGLM~\citep{DBLP:journals/corr/abs-2308-07134}, LLaGA~\citep{DBLP:conf/icml/Chen0JSW24}, ENGINE~\cite{DBLP:conf/ijcai/ZhuWST24}, SimTeG~\cite{duan2023simteg}, GraphAdapter~\cite{DBLP:conf/www/HuangHYBTCZ24}, OFA~\cite{DBLP:conf/iclr/0057FKLT0Z24}, LLM4GraphTopology (short as LLM4GT)~\cite{DBLP:journals/corr/abs-2311-14324}, GraphPrompter~\cite{DBLP:conf/www/0011H0C24}, and GraphICL~\cite{DBLP:conf/naacl/SunMFMT25} are reported according to the leaderboards (detailed in Appendix) and their papers. The results for TAPE+RevGAT, GIANT~\citep{DBLP:conf/iclr/ChienCHYZMD22}+RevGAT, GIANT+GCN, and DeBERTa~\citep{DBLP:conf/iclr/HeLGC21} are reported from ~\citep{DBLP:conf/iclr/HeB0PLH24}. Note that all the selected models are \textbf{fine-tuned} on the training set; methods focusing on transfer learning (e.g., UniGraph~\cite{DBLP:conf/kdd/HeSHH25} and ZeroG~\cite{DBLP:conf/kdd/0001WLY024}) are not included for fairness but are introduced in the related work. Mean and standard deviation over $5$ runs are reported. For text-output models, accuracy is evaluated by checking the exact matching between model's output and ground truth text.

Table~\ref{tab: comparison with SOTAs} presents a comparison between \model and SOTAs. \model consistently outperforms InstructGLM and LLaGA, \textbf{achieving new SOTA performance} among text-output node classifiers. Notably, this superior performance is achieved without modifying any LMs' architecture, demonstrating the effectiveness of our approach. Furthermore, \model exhibits \textbf{competitive performance compared to the best vector-output models}. Specifically, on Cora, Pubmed, and ogbn-arxiv datasets, \model performs closely to that of the SOTA vector-output models. Furthermore, on the ogbn-products dataset, \model surpasses the performance of the best vector-output model, TAPE.


\subsection{Ablation study}
To evaluate the contribution of each key component in \modelns, we conducted an ablation study on three crucial modules: (1) topological retrieval, (2) semantic retrieval, and (3) candidate label pruning. In this subsection, Flan-T5-small is used. The results in Table~\ref{tab: ablation study} demonstrate that each module consistently improves performance across all datasets. Notably, our analysis reveals that the relative importance of each component varies across different datasets. For instance, candidate label pruning greatly impacts performance for the Cora dataset, whereas its effect is less pronounced for the ogbn-products dataset. This variation in component importance underscores the adaptability of our approach, which can effectively accommodate diverse datasets with different characteristics.

\begin{table}[t!]
\centering
\resizebox{\linewidth}{!}{
\begin{tabular}{l|llll}
\toprule
\textbf{Model} & \textbf{Cora} & \textbf{Pubmed} & \textbf{ogbn-arxiv} & \textbf{ogbn-products} \\
\midrule
T+S     & 85.52 {\scriptsize($\downarrow$5.62)} & 94.40 {\scriptsize($\downarrow$0.40)} & 72.91 {\scriptsize($\downarrow$2.48)} & 79.83 {\scriptsize($\downarrow$1.90)} \\
T+L     & 87.27 {\scriptsize($\downarrow$3.87)} & 94.32 {\scriptsize($\downarrow$0.48)} & 73.79 {\scriptsize($\downarrow$1.60)} & 81.05 {\scriptsize($\downarrow$0.68)} \\
S+L     & 90.25 {\scriptsize($\downarrow$0.89)} & 94.26 {\scriptsize($\downarrow$0.54)} & 73.46 {\scriptsize($\downarrow$1.93)} & 79.06 {\scriptsize($\downarrow$2.67)} \\
\textbf{T+S+L} & \textbf{91.14} & \textbf{94.80} & \textbf{75.39} & \textbf{81.73} \\
\bottomrule
\end{tabular}
}
\vspace{-2mm}
\caption{Ablation study results (accuracy \%). T, S, and L denote topological retrieval, semantic retrieval, and label pruning, respectively. $\downarrow$ indicates accuracy drop compared to the full model (T+S+L).}
\label{tab: ablation study}
\end{table}

\begin{table}[t!]
\centering
\resizebox{\linewidth}{!}{
\begin{tabular}{l|llll}
\toprule
\textbf{Training} & \textbf{Cora} & \textbf{Pubmed} & \textbf{ogbn-arxiv} & \textbf{ogbn-products} \\ \midrule
Joint & 91.52 & 94.52 & 74.87 & 82.29 \\
Separate & 91.14 & 94.80 & 75.39 & 81.73 \\ \bottomrule
\end{tabular}
}
\vspace{-2mm}
\caption{Joint vs. separate training (accuracy \%).}
\label{tab: joint training}
\end{table}

\subsection{Multi-task training}
One of the key advantages of pure text-to-text instruction tuning is that a single model can be trained on multiple tasks with the same input-output format. To verify this, \model with Flan-T5-small is jointly trained on diverse datasets: Cora, Pubmed, ogbn-arxiv, and ogbn-products. The results in Table~\ref{tab: joint training} show that the jointly trained model achieves performance comparable to models trained separately on each individual dataset. We observe that on some datasets, such as Cora and ogbn-products, the jointly trained model even outperforms its dataset-specific counterparts.

These findings suggest that our approach can effectively \textbf{handle multiple graph datasets using a single model}, without incurring great performance losses compared to models trained individually. This capability is crucial for efficient model deployment when dealing with diverse graphs. In contrast, other approaches, such as InstructGLM, require the addition of a large token dictionary to accommodate all nodes in the joint dataset, which hinders their ability to achieve similar generality. Moreover, most vector-output models, including TAPE, are limited by their predefined input-output dimensions, making them inflexible and unable to handle multiple datasets.

%% file: Files/06Conclusion.tex
\section{Conclusion}
We introduce a novel framework \model for node classification on TAGs via text-to-text instruction-tuning. Our approach is built upon two key innovations: (1) topological and semantic retrieval of relevant nodes and (2) a lightweight GNN textual guidance. Extensive experimental results demonstrated (1) the effectiveness of our framework, which consistently outperformed the best text-output node classifiers while achieving performance comparable to SOTA vector-output node classifiers, and (2) the flexibility of \model, which can be jointly trained over multiple datasets without performance drop. These findings suggest a promising direction for harnessing the power of LMs in graph learning tasks.


%% file: Files/07Appendix.tex
\begin{center}
   \textbf{\LARGE Appendix} 
\end{center}

\vspace{3mm}
This appendix is organized as follows
\begin{itemize}
    \item Section~\ref{sec: intro of GNN}: introduction of the architecture and training of GNNs used in this paper.
    \item Section~\ref{sec: interpret distribution matching loss}: interpretation of the distribution matching loss.
    \item Section~\ref{sec: templates}: templates used in this paper for node classification.
    \item Section~\ref{sec: algorithm}: the training algorithm of \modelns.
    \item Section~\ref{sec: dataset statistics}: detailed dataset statistics and their leaderboards.
    \item Section~\ref{sec: hyperparameters}: hyperparameters and pretrained backbone models.
    \item Section~\ref{sec: additional exp}: additional experiments.
    \begin{itemize}
        \item Section~\ref{sec: additional efficiency exp}: additional efficiency studies.
        \item Section~\ref{sec: additional hyperparameter exp}: additional experiments on the backbone GNN selections, topological and semantic retrievers, and the PPR steps.
        \item Section~\ref{sec: additional link pred exp} additional experiments on the link prediction tasks.
    \end{itemize}
    \item Section~\ref{sec: additional related work}: extended related work
\end{itemize}

\section{Architecture and Training of the Graph Neural Network $\psi$}
\label{sec: intro of GNN}


In our setting, semi-supervised node classification problem, $\mathcal V, \mathcal E, \mathcal T, \mathcal Y_{\mathrm{train}}$ are accessible during training. Since Graph Neural Networks (GNNs)~\cite{qiu2025graph} are not inherently capable of processing textual features, a pretrained text encoder is used to generate $d$-dimensional dense embeddings for each node
\begin{align}
    \mathtt{Encoder}_{\psi_1}(t_i) = \mathbf h^{(0)}_i\in\mathbb R^{d}, \forall i\in {1,\dots, n}
    \label{eq: encoder of GNN}
\end{align}
In our implementation, the text encoder is all-MiniLM-L6-v2, a member of the Sentence Transformers. Subsequently, we apply a standard graph neural network, GraphSAGE~\citep{DBLP:conf/nips/HamiltonYL17},  whose iterative architecture is
\begin{align}
    &\mathbf h^{(l)}_i = \sigma^{(l)} \Big(\mathtt{MEAN}\Big(\mathtt{Message}_i\Big)\cdot \mathbf W^{(l)}\Big)\\
    &\mathtt{Message}_i = \{\mathbf h^{(l-1)}_i\}\cup\{\mathbf h^{(l-1)}_j: (v_i,v_j)\in\mathcal E\}
\end{align}
where $\sigma^{(l)}$ is the activation function and $\mathbf W^{(l)}\in\mathbb R^{d\times d}$ is the learnable parameter of each layer. For an $L$-layer network, in the last layer, $\sigma^{(L)}$ is $\mathtt{Softmax}$ and $\mathbf W^{(L)}\in \mathbb R^{d\times c}$ so that $\mathbf h_i^{(L)}\in \mathbb R^c$ is the prediction vector. The typical loss used for training the GNN is negative log-likelihood $\mathcal L_{\mathrm{NLL}}(\mathbf h_i^{(L)}, y_i)$ for all the nodes in the training set $\mathcal Y_{\mathrm{train}}$. The complete set of trainable parameters is denoted as $\psi=\{\psi_1\}\cup\{\mathbf W^{(l)}\}_{l=1}^L$.

\section{Interpretation of the distribution matching loss}
\label{sec: interpret distribution matching loss}

We recap the objective function. For notation brevity, we use $t_i$ to denote the input target node with its pruned candidates: $(t_{\mathrm{target}}, t_{\mathrm{candidates}})$:
\begin{align}
    \mathrm{KL}(\tilde{p}_{\mathrm{LM}}(\cdot|t_i,y_i)\|p_{\phi}(\cdot|t_i))
\end{align}
where the stop gradient operator is removed if we only compute the gradient with respect to $\phi$ and 
\begin{align}
    p_{\phi}(t_j|t_i) &= \frac{e^{s_\phi (t_i,t_j)}}{\sum_{t_k\in\mathcal D}e^{s_\phi (t_i,t_k)}}\\
    \tilde{p}_{\mathrm{LM}}(t_j|t_i, y_i) &= \frac{e^{p_{\mathrm{LM}}(y_i|t_i, t_j)}}{\sum_{k\in\mathcal N_i}e^{p_{\mathrm{LM}}(y_i|t_i, t_k)}}
\end{align}
For notation brevity, we replace $\sum_{t_k\in\mathcal D}$ with $\sum_z$ if there is no ambiguity. Then 
\begin{align}
    &\quad\ \min_\phi\ \mathrm{KL}\big(\tilde{p}_{\mathtt{LM}}(\cdot|t_i,y_i)  \| p_\phi(\cdot|t_i) \big)\\
    &\Leftrightarrow \min_\phi -\sum_z \tilde{p}_{\mathtt{LM}}(z|t_i,y_i) \log [p_\phi(z|t_i)]\\
    &= -\sum_z \tilde{p}_{\mathtt{LM}}(z|t_i,y_i) \log \left(\frac{e^{s_\phi(z,t_i)}}{\sum_{z'}e^{s_\phi(z',t_i)}}\right)\\
    &= \sum_z \tilde{p}_{\mathtt{LM}}(z|t_i,y_i)\log\left(\sum_{z'}e^{s_\phi(z',t_i)}\right)\nonumber\\
    &- \sum_z \tilde{p}_{\mathtt{LM}}(z|t_i,y_i) s_\phi(z,t_i)\\
    &= \log\left(\sum_{z}e^{s_\phi(z,t_i)}\right)\nonumber\\
    &- \sum_z \tilde{p}_{\mathtt{LM}}(z|t_i,y_i) s_\phi(z,t_i)
\end{align}
Hence,
\begin{align}
    \nabla \mathrm{KL} &= \frac{\sum_z e^{s_\phi(z,t_i)}\nabla s_\phi(z,t_i)}{\sum_{z'}e^{s_\phi(z',t_i)}})\nonumber\\
    &- \sum_z \tilde{p}_{\mathtt{LM}}(z|t_i,y_i) \nabla s_\phi(z,t_i)\\
    &= \sum_z \left(p_\phi(z|t_i)-\tilde{p}_{\mathtt{LM}}(z|t_i,y_i)\right)\nonumber\\ 
    &\cdot \nabla s_\phi(z,t_i)\\
    &= \sum_z \left(1-\frac{\tilde{p}_{\mathtt{LM}}(z|t_i,y_i)}{p_\phi(z|t_i)}\right)\nonumber\\
    &\cdot p_\phi(z|t_i)\nabla s_\phi(z,t_i)
\end{align}
After changing the notation back from $\sum_z$ to $\sum_{t_k\in\mathcal D}$, we have
\begin{align}
   \nabla \mathrm{KL} = & \sum_{t_k\in\mathcal D} \left(1-\frac{\tilde{p}_{\mathtt{LM}}(t_j|t_i,y_i)}{p_\phi(t_j|t_i)}\right)\nonumber\\
   &\cdot p_\phi(t_j|t_i)\nabla s_\phi(t_j,t_i)
\end{align}
whose rationale is that \textbf{if the LM's feedback greatly prefers the neighbor $v_j$ (and its associated text $t_j$), larger than its probability to be retrieved by the retriever (i.e., $\frac{\tilde{p}_{\mathtt{LM}}(t_j|t_i,y_i)}{p_\phi(t_j|t_i)}>1$), then the similarity score between $t_i$ and $t_j$ will increase}, i.e., improve the probability of $t_j$ to be retrieved.

\section{Templates}
\label{sec: templates}
Table~\ref{tab: templates} presents templates used in this paper. We design the "Citation" template for the Cora, Pubmed, and ogbn-arxiv datasets and the "Amazon" template for the ogbn-products dataset.

Drawing inspiration from the findings of~\cite{DBLP:conf/iclr/HeB0PLH24}, who demonstrated the efficacy of positioning the title after the main content for certain datasets, we have also introduced two additional template variations: "Citation, Title Last" and "Amazon, Title Last."

\begin{table*}[t!]
\centering
\small
\begin{tabularx}{\textwidth}{l|X}
\toprule
\textbf{Template Name} & \textbf{Prompt Text} \\ \midrule

\begin{tabular}[c]{@{}l@{}}Citation\\ (Cora, Pubmed, ogbn-arxiv)\end{tabular} &
\ttfamily Please classify the following paper into {\color{blue}\{pruned label candidates\}} based on the provided information{\color{gray}\textbackslash n}Title: {\color{blue}\{target node's title\}}{\color{gray}\textbackslash n}Content: {\color{blue}\{target node's abstract\}}{\color{gray}\textbackslash n}Related papers: {\color{blue}\{retrieved nodes' titles\}} \\ \midrule

\begin{tabular}[c]{@{}l@{}}Citation, Title Last\\ (Cora, Pubmed, ogbn-arxiv)\end{tabular} &
\ttfamily Please classify the following paper into {\color{blue}\{pruned label candidates\}} based on the provided information{\color{gray}\textbackslash n}Content: {\color{blue}\{target node's abstract\}}{\color{gray}\textbackslash n}Related papers: {\color{blue}\{retrieved nodes' titles\}}{\color{gray}\textbackslash n}Title: {\color{blue}\{target node's title\}} \\ \midrule

\begin{tabular}[c]{@{}l@{}}Amazon\\ (ogbn-products)\end{tabular} &
\ttfamily Please classify the following Amazon product into {\color{blue}\{pruned label candidates\}} based on the provided information{\color{gray}\textbackslash n}Product name: {\color{blue}\{target node's title\}}{\color{gray}\textbackslash n}Description: {\color{blue}\{target node's description\}}{\color{gray}\textbackslash n}Related products: {\color{blue}\{retrieved nodes' titles\}} \\ \midrule

\begin{tabular}[c]{@{}l@{}}Amazon, Title Last\\ (ogbn-products)\end{tabular} &
\ttfamily Please classify the following Amazon product into {\color{blue}\{pruned label candidates\}} based on the provided information{\color{gray}\textbackslash n}Description: {\color{blue}\{target node's description\}}{\color{gray}\textbackslash n}Related products: {\color{blue}\{retrieved nodes' titles\}}{\color{gray}\textbackslash n}Product name: {\color{blue}\{target node's title\}} \\

\bottomrule
\end{tabularx}
\vspace{-2mm}
\caption{Templates used for all datasets.}
\label{tab: templates}
\end{table*}

\section{Algorithm}
\label{sec: algorithm}

A step-by-step process for fine-tuning \modelns, processing one training node per step, is presented in Algorithm~\ref{alg: training}. This procedure can be readily extended to mini-batch settings.

\begin{algorithm}[t]
\caption{Training procedure for \modelns}
\label{alg: training}
\begin{algorithmic}[1]
\State \textbf{Input:} 
\Statex \hspace{1em}(1) A graph $\mathcal G = (\mathcal V, \mathcal E, \mathcal T)$ and training labels $\mathcal Y_{\mathrm{train}}$; 
\Statex \hspace{1em}(2) initialized backbone LM $\theta$; 
\Statex \hspace{1em}(3) initialized semantic encoder $\phi$;
\Statex \hspace{1em}(4) initialized GNN $\psi$.

\State \textbf{Preprocessing:}
\Statex \hspace{1em}(1) Pretrain GNN $\psi$ on $(\mathcal G, \mathcal Y_{\mathrm{train}})$.
\Statex \hspace{1em}(2) Generate prototypes and their text via Eqs.~(\ref{eq: prototype selection}) and~(\ref{eq: prototype text}).
\Statex \hspace{1em}(3) Generate pruned label candidates for each node via Eq.~(\ref{eq: pruned label candidates}).

\While{$\theta$ and $\phi$ not converged}
  \State Sample node $v_i \sim \mathcal V$ with text $t_i$.
  
  \State Retrieve relevant nodes' text $t_{\mathrm{retri},i}$ via:
  \Statex \hspace{1em}topological retrieval (Eq.~(\ref{eq: ppr selection})) and/or 
  \Statex \hspace{1em}semantic retrieval (Eq.~(\ref{eq: RAG retrieve})).
  
  \State Construct prompt with $t_i$, $t_{\mathrm{retri},i}$, and $t_{\mathrm{candidates},i}$ (from Preprocessing step (3)), based on the template (e.g., Figure~\ref{fig: our template}).

  \State Update $\theta$ based on Eq.~(\ref{eq: NLL for finetuning backbone LM}).

  \State Compute $p_{\phi}(\cdot|t_i)$ via Eq.~(\ref{eq: retrieval distribution}).

  \State Perform LM inference $|\mathcal D|$ times for:
  \Statex \hspace{1em}$\{p_{\mathrm{LM}}(y_i|t_i, t)\}_{t \in \mathcal D}$ and $\tilde{p}_{\mathrm{LM}}(\cdot|t_i, y_i)$.

  \State Update $\phi$ based on Eq.~(\ref{eq: KL loss}).
\EndWhile
\end{algorithmic}
\end{algorithm}

\section{Dataset Statistics}
\label{sec: dataset statistics}

We present the detailed statistics of datasets used in this paper in Table~\ref{tab: dataset statistics}.

All the baseline methods' performance on the Cora, Pubmed, and ogbn-arxiv is reported from the public leaderboards~\footnote{\url{https://paperswithcode.com/sota/node-classification-on-cora-60-20-20-random}}\footnote{\url{https://paperswithcode.com/sota/node-classification-on-pubmed-60-20-20-random}}\footnote{\url{https://ogb.stanford.edu/docs/leader_nodeprop/}} and their published papers.

The ogbn-products dataset used in this paper is a subset of the original ogbn-products dataset~\cite{DBLP:journals/corr/abs-2005-00687} from TAPE~\cite{DBLP:conf/iclr/HeB0PLH24}. We follow the settings in TAPE and report baseline methods' performance from the TAPE~\cite{DBLP:conf/iclr/HeB0PLH24} paper.


\begin{table*}[t!]
\centering
\renewcommand{\arraystretch}{1.1}
\resizebox{\textwidth}{!}{%
\begin{tabular}{l
                >{\raggedleft\arraybackslash}p{2.2cm} 
                >{\raggedleft\arraybackslash}p{2.4cm} 
                >{\raggedleft\arraybackslash}p{2.2cm} 
                l 
                l}
\toprule
\textbf{Name} & \textbf{\# Nodes} & \textbf{\# Edges} & \textbf{\# Classes} & \textbf{Split Strategy} & \textbf{Evaluation Metric} \\
\midrule
Cora           & \numprint{2708}     & \numprint{10556}     & 7   & Random 60/20/20\% & Accuracy \\
Pubmed         & \numprint{19717}    & \numprint{88648}     & 3   & Random 60/20/20\% & Accuracy \\
ogbn-arxiv     & \numprint{169343}   & \numprint{1166243}   & 40  & Given split        & Accuracy \\
ogbn-products  & \numprint{54025}    & \numprint{198663}    & 47  & Given split        & Accuracy \\
\bottomrule
\end{tabular}%
}
\vspace{-2mm}
\caption{Dataset statistics.}
\label{tab: dataset statistics}
\end{table*}

\section{Selected Hyperparameters}
\label{sec: hyperparameters}

We report the hyperparameters for every dataset in Table~\ref{tab: hyperparameter selection}. As mentioned in the main content, we use two pretrained all-MiniLM-L6-v2 models as the dual encoder and the Flan-T5-small/base/large models as the backbone; they are all publicly available\footnote{\url{https://huggingface.co/sentence-transformers/all-MiniLM-L6-v2}}\footnote{\url{https://huggingface.co/docs/transformers/en/model_doc/flan-t5}}. More detailed hyperparameters will be released with the code upon publication.


\begin{table*}[t!]
\centering
\renewcommand{\arraystretch}{1.1}
\begin{tabular}{l|cccc}
\toprule
\textbf{Hyperparameter} & \textbf{Cora} & \textbf{Pubmed} & \textbf{ogbn-arxiv} & \textbf{ogbn-products} \\
\midrule
\# PPR neighbors           & 5     & 2     & 5     & 5 \\
\# Semantic neighbors      & 5     & 2     & 5     & 5 \\
Prompt template            & Citation & Citation & Citation, Title Last & Amazon \\
\# Candidate labels        & 3     & 2     & 3     & 3 \\
LM learning rate           & $1 \times 10^{-4}$ & $1 \times 10^{-4}$ & $1 \times 10^{-4}$ & $1 \times 10^{-4}$ \\
Retriever learning rate    & $1 \times 10^{-5}$ & $1 \times 10^{-5}$ & $1 \times 10^{-5}$ & $1 \times 10^{-5}$ \\
Weight decay               & 0     & 0     & 0     & 0 \\
\bottomrule
\end{tabular}
\vspace{-2mm}
\caption{Selected hyperparameters for \modelns{} across different datasets.}
\label{tab: hyperparameter selection}
\end{table*}

\section{Additional Experiments}
\label{sec: additional exp}

\subsection{Additional efficiency study}
\label{sec: additional efficiency exp}




\paragraph{Memory usage.} Memory usage is linear concerning batch size. We report the memory usage of \model with different backbone LMs in Table~\ref{tab: memory usage}, where we set the batch size to $1$ and we found the experimental results reasonable because more powerful backbone LMs require more GPU memory.

\begin{table}[h!]
\centering
\begin{tabular}{l|r}
\toprule
\textbf{Model} & \textbf{Memory} \\\midrule
\model (T5-small) & \numprint{3098} \\
\model (T5-base) & \numprint{6572} \\
\model (T5-large) & \numprint{20308}\\\bottomrule
\end{tabular}%
\vspace{-2mm}
\caption{GPU Memory usage (MB) with different LMs.}
\label{tab: memory usage}
\end{table}

\begin{figure}[t!]
\centering
\includegraphics[width=.45\textwidth]{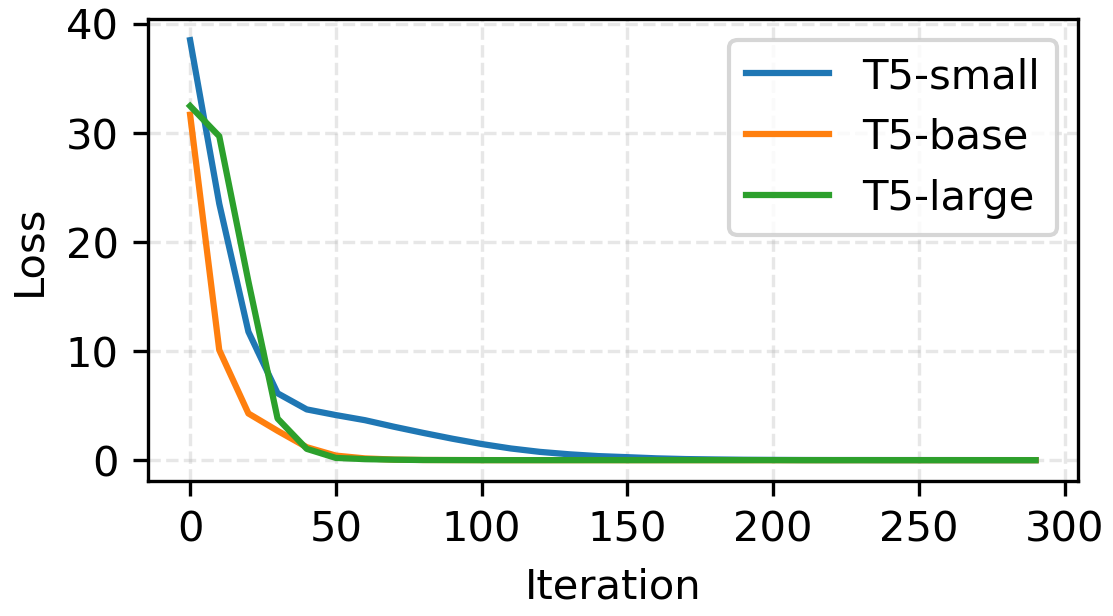}
\caption{Convergence curve of \modelns.}
\label{fig: convergence curve}
\end{figure}

\paragraph{Convergence curve.} We train \model with different backbone LMs: FLAN-T5-small/base/large on the Cora dataset and plot their loss curves regarding updating steps in Figure~\ref{fig: convergence curve}. In this experiment, the batch size is $16$. It shows that our proposed \model converges smoothly and quickly when equipped with various LMs of different scales.

\begin{table}[t!]
\centering
\begin{tabular}{l|r}
\toprule
\textbf{Module} & \textbf{FLOPs ($10^9$)} \\ \midrule
Retriever & 2.3 \\
T5-small & 71.7 \\
T5-base & 257.2 \\
T5-large & 845.4 \\ \bottomrule
\end{tabular}%
\vspace{-2mm}
\caption{FLOPs comparison between different modules.}
\label{tab: flops}
\end{table}

\paragraph{FLOPs.}
The floating point operations (FLOPs) of \model are studied. Specifically, the computation of our \model includes (1) precomputing PPR neighbors, (2) training and inference of the semantic retriever $\phi$, and (3) training and inference of the LM $\theta$. Hence, the extra on-the-fly computation cost is from the semantic retriever $\phi$ (all-MiniLM-L6-v2 in our experiments). We report the FLOPs of the retriever and different LM backbones in Table~\ref{tab: flops}. The results show that (1) the retriever only adds a tiny amount of FLOPs to the backbone LMs and (2) our proposed \model is efficient.

\paragraph{Running time.} The running time (both forward and backpropagation) of the semantic retriever and the backbone LMs on the Cora dataset is recorded. The batch size is $1$. This experiment is tested on an NVIDIA A100-SXM4-40GB. Table~\ref{tab: running time} shows that the semantic retriever only adds very limited on-the-fly computation overhead compared to the LM, showing the efficiency of \modelns.

\begin{table}[t!]
\centering
\begin{tabular}{l|rr}
\toprule
\textbf{Module} & \textbf{Forward} & \textbf{Backprop} \\ \midrule
Retriever & 14.7 & 6.1 \\
T5-small & 90.0 & 32.0 \\
T5-base & 104.4 & 66.6 \\
T5-large & 277.2 & 197.0 \\ \bottomrule
\end{tabular}%
\vspace{-2mm}
\caption{Running time (ms) of different modules.}
\label{tab: running time}
\end{table}






\subsection{Additional hyperparameter study}
\label{sec: additional hyperparameter exp}
In this section, we study the model's performance with various hyperparameters. 

\paragraph{Selection of the backbone GNN.}

Specifically, we study the performance of \model equipped with different GNNs. we compared the performance of \model equipped with GraphSAGE (used in the reported results) with the counterpart equipped with GCN~\cite{DBLP:conf/iclr/KipfW17}. The comparison is in Table~\ref{tab: parameter study with different GNNs}.

\begin{table*}[h!]
\centering
\begin{tabular}{l|llll}
\toprule
\textbf{Model} & \textbf{Cora} & \textbf{Pubmed} & \textbf{ogbn-arxiv} & \textbf{ogbn-products} \\ \midrule
GraphSAGE & 91.14 & 94.80 & 75.39 & 81.73 \\
GCN & 90.98 & 94.85 & 75.21 & 81.82 \\ \bottomrule
\end{tabular}%
\vspace{-2mm}
\caption{Performance (accuracy \%) comparison of \model equipped with different GNNs.}
\label{tab: parameter study with different GNNs}
\end{table*}

\begin{table}[h!]
\centering
\begin{tabular}{c|c}
\toprule
\textbf{\# Neighbors} & \textbf{Accuracy (\%)} \\
\midrule
1   & 75.18 \\
3   & 75.76 \\
5   & 75.39 \\
7   & 75.19 \\
9   & 76.05 \\
10  & \textbf{76.45} \\
15  & 75.99 \\
20  & 74.81 \\
25  & 74.48 \\
\bottomrule
\end{tabular}
\vspace{-2mm}
\caption{Accuracy (\%) of \model{}on ogbn-arxiv with different numbers of PPR-retrieved neighbors. The best result is \textbf{bolded}.}
\label{tab: parameter study with different PPR neighbors}
\end{table}

We observed that the performance is nearly identical between GCN and GraphSAGE. This can be attributed to two factors: (1) the classification performances of GCN and GraphSAGE are similar, and (2) the GNN is used to generate prototypes and prune candidate labels, which \textbf{does not require a highly powerful GNN} for accurate classification.

\paragraph{Number of PPR retrieved nodes.}

Next, we examined the relationship between the model performance and the number of nodes retrieved. In this auxiliary experiment, we fixed the number of nodes retrieved by semantic retrieval at $5$ and varied the number of nodes retrieved by PPR retrieval. The results are reported in Table~\ref{tab: parameter study with different PPR neighbors}


\begin{table*}[t]
\centering
\begin{tabular}{l|llll}
\toprule
\textbf{Retrieval Technique} & \textbf{Cora} & \textbf{Pubmed} & \textbf{ogbn-arxiv} & \textbf{ogbn-products} \\
\midrule
$1$-hop neighbors & 90.59 & 94.33 & 73.97 & 79.53 \\
GAE               & 90.83 & 94.42 & 74.01 & 79.85 \\
PPR neighbors     & \textbf{91.14} & \textbf{94.80} & \textbf{75.39} & \textbf{81.73} \\
\bottomrule
\end{tabular}
\vspace{-2mm}
\caption{Accuracy (\%) of \model{}with different \emph{topological} retrieval techniques across datasets. The best result for each dataset is \textbf{bolded}.}
\label{tab: parameter study with different topological retrieval techniques}
\end{table*}


\begin{table*}[t]
\centering
\begin{tabular}{l|llll}
\toprule
\textbf{Retrieval Technique} & \textbf{Cora} & \textbf{Pubmed} & \textbf{ogbn-arxiv} & \textbf{ogbn-products} \\
\midrule
Simple semantic retriever & 90.68 & 94.37 & 74.46 & 81.21 \\
SimTeG-tuned simple retriever & --- & --- & 74.70 & --- \\
Prototype-based retriever (ours) & \textbf{91.14} & \textbf{94.80} & \textbf{75.39} & \textbf{81.73} \\
\bottomrule
\end{tabular}
\vspace{-2mm}
\caption{Accuracy (\%) of \model{}with different \emph{semantic} retrieval techniques across datasets. The best result for each dataset is \textbf{bolded}.}
\label{tab: parameter study with different semantic retrieval techniques}
\end{table*}


Interestingly, we found that the model's performance remains relatively stable when the number of PPR nodes is less than 15. However, the performance degrades when too many nodes are retrieved (more than 15). A possible explanation is that when the number of PPR nodes becomes too large, every target node's \textbf{retrieved nodes become similar} (e.g., some hub nodes are retrieved by most nodes), reducing the discriminativeness of each target node. This phenomenon is reminiscent of the "oversmoothing" problem~\cite{DBLP:conf/aaai/LiHW18} in GNNs, where a GNN with too many layers and a large receptive field produces indistinguishable latent representations for all the nodes.

\paragraph{Other topological retrieval options.}

In this auxiliary experiment, we use the link predictor to retrieve relevant neighbors. Specifically, we trained a graph autoencoder (GAE)~\cite{DBLP:journals/corr/KipfW16a}, a basic graph neural network-based link predictor, on the given graph. Then, we retrieved the \textbf{top-$5$ most confident neighbors from the reconstructed graph} to replace those obtained through PPR retrieval. The results are presented in Table~\ref{tab: parameter study with different topological retrieval techniques}, where Flan-T5-small is used as the backbone LM. For better reference, we also provide a version where PPR retrieval is replaced with retrieving from $1$-hop neighbors.

We observe that both $1$-hop neighbor retrieval and GAE perform worse than their PPR counterparts. A possible reason is that both 1-hop neighbor retrieval and GAE are \textbf{local} retrieval methods, whereas PPR can effectively capture the \textbf{global} structure. Additionally, we note that GAE is trained using a reconstruction loss, which means it tends to assign high confidence to \textbf{existing edges}. In other words, the neighbors retrieved by GAE would be similar to those obtained through $1$-hop neighbor retrieval, except for some low-degree nodes.

\paragraph{Other semantic retrieval options.}
This additional experiment uses different semantic retrievers to replace the prototype-based semantic retriever used in the proposed \modelns. In detail, the prototype-based semantic retrieval module is replaced with a simple semantic retriever that \textbf{selects the most textually similar nodes} via inner product. Concretely, we use two pretrained models, (1) the original all-MiniLM-L6-v2\footnote{\url{https://huggingface.co/sentence-transformers/all-MiniLM-L6-v2}} and (2) a fine-tuned all-MiniLM-L6-v2 by SimTeG~\cite{duan2023simteg}\footnote{\url{https://huggingface.co/datasets/vermouthdky/SimTeG/tree/main/ogbn-arxiv/all-MiniLM-L6-v2/main/cached_embs}}. The remaining modules, including topological retrieval and classifier guidance, were left intact, and FLAN-T5-small is used as the LM backbone. The results are reported in Table~\ref{tab: parameter study with different semantic retrieval techniques}.

We observe that the proposed prototype-based retriever is better than both the original all-MiniLM-L6-v2-based retriever and the SimTeg-tuned simple retriever. This is because:
\begin{enumerate}
    \item The training objective of the SimTeG-tuned retriever is to align the classification loss with a GNN model~\cite{duan2023simteg}, similar to knowledge distillation~\cite{DBLP:journals/corr/HintonVD15}. In other words, \textbf{the SimTeG-tuned retriever is a mixture of topological and semantic retrieval}, as the GNN incorporates both topology and node features. This means that its role partially overlaps with that of the topological PPR retriever.
    
    \item Our prototype-based retriever can retrieve textual features from \textbf{multiple nodes}, but the other two cannot achieve this.
\end{enumerate}

\paragraph{Other backbone LM options.}
We studies the other LM options beyond the T5 family; in detail, we use Llama-3.2-3B~\cite{grattafiori2024llama} finetuned with rank-8 LoRA. Results compared with the best text-output models are in Table~\ref{tab: compared with other LMs},

\begin{table*}[t!]
\centering
\begin{tabular}{l|llll}
\toprule
\textbf{Backbone LM} & \textbf{Cora} & \textbf{Pubmed} & \textbf{ogbn-arxiv} & \textbf{ogbn-products} \\ \midrule
T5-base & 91.24 & 95.03 & 76.80 & 81.91 \\
T5-large & \textbf{91.51} & 95.16 & 76.00 & 82.90 \\ 
Llama-3.2-3B & 91.45 & \textbf{95.20} & \textbf{77.12} & \textbf{83.02} \\ \bottomrule
\end{tabular}
\vspace{-2mm}
\caption{Accuracy (\%) of \model{}with different LM backbones.}
\label{tab: compared with other LMs}
\end{table*}

Overall, Llama-3.2-3B achieves performance comparable to previous backbones on Cora and Pubmed, and improvements on ogbn-arxiv and ogbn-products. The similar results on the smaller datasets (Cora and Pubmed) likely reflect their limited data scale, whereas the gains on the larger benchmarks (ogbn-arxiv and ogbn-products) suggest advantages from scaling up the LM. While we did not explore optimal configurations, we expect that further tuning could yield even larger improvements on these large-scale datasets.

\subsection{Additional link prediction experiments}
\label{sec: additional link pred exp}
The main task of this paper is on the node classification task, but we conducted a \emph{preliminary} experiment to adapt our proposed \model to the link prediction task~\cite{DBLP:conf/cikm/FuXLTH20,DBLP:conf/nips/BanZLQFKTH24}, further showcasing the generality of the proposed \modelns. A systematic study to adapt \model to link prediction tasks is interesting, and we leave it as future work.

Link prediction can be viewed as \textbf{a classification task for a pair of nodes}. For all modules, we made the following adaptations:
\begin{enumerate}
    \item We retained the topological PPR retrieval for the input node pair.
    \item We concatenated the text of the node pair as input for the semantic retriever. The prototypes used as the corpus of the semantic retriever were still generated by a pre-trained GNN, which is consistent with our approach for the node classification task.
    \item For classifier guidance, we utilized a pre-trained graph autoencoder (GAE), whose output is the connection probability for every node pair. We transformed the connection probability into plain language based on the following rules: (1) less than 0.2: "improbable", (2) 0.2 to 0.4: "unlikely", (3) 0.4 to 0.6: "maybe", (4) 0.6 to 0.8: "likely", and (5) more than 0.8: "highly likely". The GAE's prediction (in plain language) was then incorporated into the following template.
    \item The template is in the following format:
\end{enumerate}

\begin{tcolorbox}[
    colback=gray!10,
    colframe=gray!60,
    title=Prompt Template for Link Prediction,
    fonttitle=\bfseries,
    boxrule=0.5pt,
    rounded corners,
    breakable,
    enhanced
]
\small\ttfamily
Please determine if the following two papers are related or not.\\

Paper 1's title: {\color{blue}\{Paper 1's title\}}\\
Paper 1's abstract: {\color{blue}\{Paper 1's abstract\}}\\
Paper 1's related works: {\color{blue}\{Paper 1's PPR neighbors' titles\}}\\

Paper 2's title: {\color{blue}\{Paper 2's title\}}\\
Paper 2's abstract: {\color{blue}\{Paper 2's abstract\}}\\
Paper 2's related works: {\color{blue}\{Paper 2's PPR neighbors' titles\}}\\

Other related works: {\color{blue}\{Semantic retrieved nodes' titles\}}\\

An expert link prediction model predicted that the possibility of these two papers being related is: {\color{blue}\{GAE's prediction\}}\\

Do you think these two papers are related or not?\\
Please answer Yes or No.
\end{tcolorbox}

We conducted preliminary experiments on the Cora dataset, following the settings from the benchmark\footnote{\url{https://paperswithcode.com/paper/variational-graph-auto-encoders}}. In this setup, 5\% and 10\% of edges were removed for validation and testing, respectively. Also, an equal number of non-connected node pairs were used as negative samples. The accuracy results are reported in the following table.

\begin{table}[t]
\centering
\begin{tabular}{l|l}
\toprule
\textbf{Method} & \textbf{Accuracy} \\
\midrule
GAE & 89.29 \\
\model{} (T5-small) & 93.59 \\
\model{} (T5-base)  & \textbf{94.25} \\
\bottomrule
\end{tabular}
\vspace{-2mm}
\caption{Accuracy (\%) on the preliminary link prediction task for the Cora dataset.}
\label{tab: link prediction results}
\end{table}


Our key findings are as follows:
\begin{enumerate}
    \item Our proposed \model can indeed be effectively adapted to link prediction tasks.
    \item By leveraging a classic link predictor (GAE), our \model achieves a significant performance boost over the backbone predictor GAE, which aligns with our observations in node classification tasks.
\end{enumerate}


\section{Extended Related Work}
\label{sec: additional related work}

\noindent\textbf{LMs for graphs}. Recent studies have explored the ability of LMs to understand graph topology by investigating problems such as substructure recall~\citep{DBLP:journals/corr/abs-2402-11821}, connectivity~\citep{DBLP:conf/nips/WangFHTHT23,DBLP:journals/corr/abs-2402-05862}, node/edge counting~\citep{DBLP:journals/corr/abs-2402-05862}, and spatial-temporal graphs~\citep{DBLP:conf/kdd/ZhangWZ0Q024}. These efforts demonstrate that, although LMs are not trained explicitly for graph tasks, they can exhibit non-trivial reasoning abilities with suitable prompts/encodings.

Building on these findings, more work has explored the use of LMs for tasks on text-attributed graphs (TAGs). For node classification, several methods directly prompt LMs with node and neighborhood textual information to perform inference~\citep{DBLP:journals/corr/abs-2308-07134,DBLP:journals/corr/abs-2310-01089,DBLP:journals/corr/abs-2402-03720,DBLP:journals/corr/abs-2310-18152,DBLP:conf/acl/0002WCDZX025,DBLP:journals/corr/abs-2311-14324}. These works vary in how they construct the input prompts, such as using structured templates~\citep{DBLP:journals/corr/abs-2311-14324} or similarity-based neighbor selection~\citep{DBLP:journals/corr/abs-2402-03720}, and have shown promising performance.

Beyond node classification, LMs have also been adapted for link prediction on TAGs~\citep{DBLP:journals/corr/abs-2305-14321, DBLP:journals/corr/abs-2403-04780,DBLP:conf/iclr/0057FKLT0Z24,DBLP:conf/www/0011H0C24}, where the textual features of node pairs and their neighborhoods are used to determine potential links. These approaches often incorporate task-specific prompt engineering or graph-aware context construction to maximize LM performance.

A notable direction is transfer and zero-shot learning, where foundation models are trained to generalize across multiple graphs or tasks. Several methods aim to build unified models with in-context learning or instruction tuning capabilities~\citep{DBLP:conf/sigir/Tang00SSCY024,DBLP:conf/naacl/SunMFMT25,DBLP:conf/cikm/Pan00H024,DBLP:conf/kdd/HeSHH25,DBLP:journals/corr/abs-2408-10700,DBLP:conf/kdd/0001WLY024}. These include approaches that pretrain on diverse graph datasets using graph-text alignment objectives~\citep{DBLP:conf/kdd/0001WLY024,zhu2025llm}, instruction tuning on graph-related tasks~\citep{DBLP:conf/naacl/SunMFMT25}, or modular architectures for generalization~\citep{DBLP:conf/kdd/HeSHH25}. Notably, GraphCLIP~\citep{zhu2025graphclip} AnyGraph~\citep{DBLP:journals/corr/abs-2408-10700}, and RiemannGFM~\cite{sun2025riemanngfm} illustrate how cross-domain graph pretraining with LMs can lead to strong zero-shot or few-shot transfer.

Several studies also investigate graph reasoning with LMs~\cite{yan2025answer,DBLP:journals/sigkdd/LiuWT25}. Recent work explores the use of chain-of-thought prompting~\citep{DBLP:conf/acl/JinXZRZL0TWM024,DBLP:conf/iclr/Qiu0BT25} and tool-augmented querying~\cite{zhang2023graph}, which equip LMs with step-by-step planning capabilities to solve complex graph tasks.

Finally, hybrid models that combine LMs and GNNs have also been proposed. GIANT~\citep{DBLP:conf/iclr/ChienCHYZMD22} and GLEM~\citep{DBLP:conf/iclr/0002QLYL00023} encode textual features in a graph-aware manner to jointly leverage language and structure. TAPE~\citep{DBLP:conf/iclr/HeB0PLH24} augments the node textual features using LMs before applying a GNN, demonstrating that LMs can enrich node representations for downstream prediction. Other hybrid methods explore LLM-to-GNN distillation~\citep{DBLP:conf/cikm/Pan00H024} or GNN adapters~\citep{DBLP:conf/www/HuangHYBTCZ24}, showing that language models can either enhance or transfer knowledge to traditional graph models.

In addition to the above, as this is a rapidly evolving topic, several comprehensive reviews have been proposed. Interested readers are referred to~\cite{liu2025graph,wang2025graph,ren2024survey,li2024survey,jin2024large}

\noindent\textbf{Retrieval-augmented generation (RAG)}.
Retrieval-augmented generation (RAG) enhances language models by granting access to external knowledge sources, typically by retrieving relevant documents from a large corpus and conditioning generation on them~\citep{DBLP:conf/nips/LewisPPPKGKLYR020, DBLP:conf/emnlp/KarpukhinOMLWEC20}. This approach addresses limitations of parametric models in retaining up-to-date or factual knowledge~\citep{DBLP:conf/nips/HashimotoGOL18}. Advances such as REALM~\citep{DBLP:conf/icml/GuuLTPC20} introduced end-to-end retriever-generator training. In contrast, RETRO~\citep{DBLP:conf/icml/BorgeaudMHCRM0L22} scaled RAG to large corpora using a frozen retriever and cross-attention over retrieved chunks. Later, methods like REPLUG~\citep{DBLP:journals/corr/abs-2301-12652} allowed retriever optimization even with black-box LMs. Other extensions include Fusion-in-Decoder~\citep{DBLP:conf/eacl/IzacardG21}, Atlas~\cite{izacard2023atlas}, and multimodal RAG~\citep{DBLP:conf/icml/YasunagaAS0LLLZ23,zhao2023retrieving}, which retrieves both text and images. This topic has attracted much attention; interested readers are referred to~\cite{gao2023retrieval,zhao2024retrieval} for comprehensive surveys.

Recent studies have extended retrieval-augmented generation (RAG) to graph data. Several works focus on improving graph-based retrieval: GNN-RAG~\cite{mavromatis2024gnn} leverages a GNN to retrieve reasoning paths from knowledge graphs, while GRAG~\cite{DBLP:conf/naacl/HuLZPLZ25} encodes $k$-hop ego graphs as dense embeddings to support structure-aware retrieval. HippoRAG~\cite{jimenez2024hipporag} constructs a long-term memory as a graph of fact triples and applies PPR to retrieve a contextual subgraph for generation. Other approaches, such as ATLANTIC~\cite{munikoti2023atlantic} and KnowledGPT~\cite{wang2023knowledgpt}, augment retrieval with heterogeneous graph structures or programmatic access to knowledge bases, enabling more informed and faithful generation.

In addition to improving retrieval, some works apply GraphRAG to domain-specific tasks. For example, GrapeQA~\cite{taunk2023grapeqa} and HamQA~\cite{dong2023hierarchy} focus on enhancing commonsense and multi-hop QA via graph pruning, augmentation, or hyperbolic representation. In biomedical and scientific domains,~\cite{delile2024graph} and DALK~\cite{li2024dalk} retrieve relational subgraphs from tailored knowledge graphs to improve factual QA. GraphRAG has also been applied to generation tasks:~\cite{edge2024local} perform query-focused summarization by segmenting documents into graph-based communities. For a more comprehensive introduction, readers are referred to~\cite{peng2024graph,zhu2025graph,zhang2025survey}.